\documentclass[10pt]{article}

\usepackage[letterpaper,margin=1in]{geometry}
\usepackage{times}
\usepackage{natbib}
\usepackage{amsmath,amsfonts,amssymb,bm}
\usepackage{graphicx}
\usepackage{booktabs}
\usepackage[section]{placeins}
\usepackage{float}
\usepackage{listings}
\usepackage{CJKutf8}
\usepackage[hypertexnames=false,hidelinks]{hyperref}
\usepackage{url}
\usepackage{amsmath,amsfonts,bm}

\def\eqref#1{equation~\ref{#1}}
\def\1{\bm{1}}

\DeclareMathAlphabet{\mathsfit}{\encodingdefault}{\sfdefault}{m}{sl}
\SetMathAlphabet{\mathsfit}{bold}{\encodingdefault}{\sfdefault}{bx}{n}

\title{Replay the Curvature: Accurate and Scalable NVFP4 Quantization for Large Language Model Inference}

\newcommand{\corresponding}{\textsuperscript{\ensuremath{\dagger}}}
\author{%
\parbox{0.98\textwidth}{\centering
Ruiyi Ding\textsuperscript{1}\thanks{Work done during internship at Kling AI Infra},
Jie Li\textsuperscript{1},
Kang He\textsuperscript{1},
Ziyan Liu\textsuperscript{5},\\
Chengru Song\textsuperscript{1}\corresponding,
Yuedong Xu\textsuperscript{2,3}\corresponding,
Yuan Cheng\textsuperscript{2,3,4}\corresponding\\[0.75em]
\small
\textsuperscript{1}KlingAI Research\\
\textsuperscript{2}Fudan University\quad
\textsuperscript{3}Shanghai AI Incubation and Innovation Center\\
\textsuperscript{4}Shanghai Academy of AI for Science\quad
\textsuperscript{5}University of Science and Technology of China\\
\textsuperscript{\ensuremath{\dagger}}Corresponding authors\\
\texttt{dingruiyi@klingai.com},
\texttt{songchengru@klingai.com},
\texttt{cheng\_yuan@fudan.edu.cn}
}}
\date{}

\begin{document}

\maketitle

\begin{abstract}
Large language models make weight storage and memory traffic major inference costs, motivating low-precision formats that represent each weight with only a few bits. Such formats use a scale to map floating-point values into a small codebook; NVFP4 improves local range utilization by letting every 16 E2M1 weights share an E4M3 block scale. Choosing that scale is difficult in GPTQ because quantizing one column updates those that follow, so evaluating a block independently can misestimate its final reconstruction error. Large models pose a second challenge: full-precision weights, calibration activations, and second-order state cannot all remain on one accelerator, while assigning complete layers to devices leaves each time-consuming layer solve serial. We introduce \emph{Schur Replay}, a scale-selection algorithm that reproduces the GPTQ updates caused by each block scale and scores the resulting block error after accounting for compensation from unquantized columns. Separately, our execution infrastructure keeps only the active layer resident, tiers activations across device, host, and disk, retires full-precision layers after export, and distributes independent output rows across tensor-parallel ranks. Together, the algorithm and infrastructure attain $99.35\%$ and $100.84\%$ question-weighted recovery from BF16 across seven benchmarks on Qwen3.5-397B-A17B and Llama-3.3-70B-Instruct. On the 397B model, the infrastructure reduces measured per-layer time by $15.17\times$ over ModelOpt and $23.14\times$ over LLM Compressor, with lower memory used per GPU.
\end{abstract}

\section{Introduction}
Modern large language models contain tens or hundreds of billions of parameters, making weight storage and memory traffic major costs during inference. Low-precision formats reduce the bits used by each weight, allowing larger models to fit in accelerator memory and reducing the data moved for matrix multiplication. To map a floating-point weight $w$ into a small set of representable values, quantization first chooses a scale $s$, rounds the normalized value $w/s$ to the nearest low-precision code, and approximately reconstructs the weight as the code times $s$. A common initialization sets $s$ from the largest absolute weight so that the codebook covers the observed range. With one scale for an entire tensor, however, a few large values can leave most weights using only a small fraction of that codebook. Microscaling addresses this mismatch by assigning local scales to small blocks. NVFP4 adopts this design: a tensor-wide FP32 scale sets the global range, while every 16 E2M1 weights share an E4M3 block scale, yielding approximately 4.5 bits per value \citep{rouhani2023microscaling,nvidia2025nvfp4}. This layout is efficient on modern accelerators, but makes the 16 quantization decisions interdependent: one poorly chosen block scale can distort the whole block. Post-training quantization (PTQ) must choose these values from a small calibration set without retraining, and GPTQ uses activation-derived second-order information to compensate each rounding error \citep{frantar2023gptq}.

The difficulty is that GPTQ is sequential. After quantizing one column, it updates the columns that follow to compensate for the incurred error. Two block scales that initially look similar can therefore create different later states and different final reconstruction errors. Existing NVFP4 methods improve the representation or its initialization: Four Over Six adapts the effective E2M1 range; SOAR and ScaleSweep optimize or search scales; MR-GPTQ rotates local blocks; and ARCQuant adds residual channels \citep{cook2026fouroversix,bao2026soar,lin2026scalesweep,egiazarian2026mrgptq,meng2026arcquant}. Our focus is complementary: evaluate each exportable scale in the GPTQ state that it actually creates, rather than only on the original block.

Scaling this reconstruction to hundreds of billions of parameters creates two separate execution bottlenecks. First, full-precision weights, calibration activations, and second-order state cannot all remain in accelerator memory. Second, distributing samples, experts, or complete modules does not accelerate the reconstruction of one exceptionally large layer. Public toolchains support layer-wise loading, disk offloading, and multi-device execution \citep{nvidiamodelopt,llmcompressor2024,gong2024llmc,qubitium2024gptqmodel}, but their documented designs neither realize paralleled solve of each weight matrix across tp ranks, nor make use of a strict active-layer residency and a layer-scoped activation lifecycle.

We address the algorithmic bottleneck with \emph{Schur Replay}, a group-scale selection algorithm. For each hardware-valid scale, it performs the GPTQ updates that the candidate would trigger, then scores the resulting block error after accounting for the compensation still available from unquantized columns. We also formalizes this algorithm through recurrence-exact replay and a conditional Schur objective. We address the execution bottleneck with a infrastructure framework: because different output rows follow independent GPTQ recurrences, it assigns rows of the same layer to different ranks, keeps only the active weights on device, tiers activations across device, host, and disk, and retires full-precision layers after recoverable NVFP4 export.

Our main contributions are:

\textbf{Recurrence-aware NVFP4 reconstruction.} Schur Replay evaluates each exportable scale in the sequential GPTQ state that it creates, capturing feedback within the 16-value block.

\textbf{Memory-bounded distributed execution.} Active-layer residency and tiered activation storage bound device memory, while output-row parallelism engages all tensor-parallel ranks in the current layer.

\textbf{Large-model evidence.} Combining Schur Replay with our execution infrastructure recovers $99.35\%$/$100.84\%$ of BF16 across seven benchmarks on Qwen3.5-397B-A17B/Llama-3.3-70B-Instruct. On 397B, the infrastructure makes measured per-layer compute $15.17\times$/$23.14\times$ faster than ModelOpt/LLM Compressor at 35.0\,GB peak memory per GPU; full-model totals remain projections.

\section{Related Work}
\label{sec:related-work}

\textbf{Quantization methods.} PTQ converts models without retraining: AdaRound learns rounding \citep{nagel2020adaround}; OBC/OBQ and GPTQ use second-order compensation \citep{frantar2022obc,frantar2023gptq}; SmoothQuant, AWQ, and OmniQuant calibrate distributions \citep{xiao2023smoothquant,lin2024awq,shao2024omniquant}; and QuIP\# and SpinQuant apply incoherence processing or rotations \citep{tseng2024quipsharp,liu2025spinquant}. QAT instead adapts parameters to simulated quantization noise but requires more data, compute, and memory \citep{bengio2013ste,liu2023llmqat,chen2025efficientqat,bondarenko2024lrqat}.

\textbf{NVFP4.} Four Over Six adapts the E2M1 range \citep{nvidia2025nvfp4}; SOAR and ScaleSweep optimize or search scales \citep{cook2026fouroversix,bao2026soar}; MR-GPTQ adds micro-rotations and reordering; and ARCQuant adds residual channels \citep{rouhani2023microscaling,lin2026scalesweep,egiazarian2026mrgptq,meng2026arcquant}. Native pretraining and quantization-aware distillation are training-based alternatives \citep{nvidia2025nvfp4,xin2026nvfp4qad}. These approaches change initialization or representation; Schur Replay instead scores each scale in its induced GPTQ recurrence.

\textbf{Infrastructure.} Existing stacks provide calibration, offload, framework integration, and deployment support \citep{nvidiamodelopt,llmcompressor2024,gong2024llmc,amdquark,intelneuralcompressor,or2025torchao,microsoftolive,qubitium2024gptqmodel}. LLM Compressor's documented distributed GPTQ assigns each module to one owner rank; we found no public design fullyoutilize the layerwise and parallel feature for efficient PTQ solving. 
\section{Preliminaries}
\label{sec:preliminaries}

\paragraph{Layer reconstruction.}
For a linear layer $W\in\mathbb{R}^{M\times N}$ and calibration inputs $X\in\mathbb{R}^{N\times T}$, where $M$, $N$, and $T$ denote the output width, input width, and number of calibration positions, reconstruction-based PTQ solves
\begin{equation}
    \widehat W=\operatorname*{arg\,min}_{W_q\in\mathcal Q}
    \lVert WX-W_qX\rVert_F^2.
    \label{eq:prelim-layer-reconstruction}
\end{equation}
The $\mathcal Q$ is the set of weight matrices representable by the target quantization format. Writing $\Delta=W_q-W$, $H=XX^\top$, and $\delta_m=\Delta_{m,:}^\top\in\mathbb{R}^{N}$ we obtain
\begin{equation}
    \lVert\Delta X\rVert_F^2
    =\operatorname{tr}(\Delta H\Delta^\top)
    =\sum_{m=1}^{M}\delta_m^\top H\delta_m.
    \label{eq:prelim-hessian-form}
\end{equation}
Thus all output rows share one input Hessian but contribute independent quadratic objectives. OBS and OBC/OBQ exploit this structure by compensating each forced quantization error through the remaining unconstrained coordinates \citep{hassibi1993obs,frantar2022obc}.

\paragraph{GPTQ recurrence.}
GPTQ scales this compensation to transformer layers by quantizing every output row in a common input-column order and reusing one Hessian factor \citep{frantar2023gptq}. Let $\pi$ be the chosen column permutation (the identity unless activation ordering is enabled). Applying $\pi$ to the columns of $W$ and the corresponding rows of $X$ preserves $H=XX^\top$ in the permuted coordinate system; henceforth $W$, $X$, $H$, and all column indices refer to that order. After damping and handling structurally zero columns, let
\begin{equation}
    H^{-1}=U^\top U,
    \label{eq:prelim-gptq-factor}
\end{equation}
where $U$ is upper triangular and $u_{ii}=U_{ii}$. For a working row $w$ in this order and column $i$, GPTQ quantizes the current compensated value to $q_i$ and updates
\begin{equation}
    e_i=\frac{w_i-q_i}{u_{ii}},
    \qquad
    w_{i:}\leftarrow w_{i:}-e_iU_{i,i:}.
    \label{eq:prelim-gptq-recurrence}
\end{equation}
The recurrence is sequential across input columns because each error changes later values, whereas output rows remain independent. NVFP4 shares one scale across consecutive columns, so selecting that scale from the group's initial weights ignores the values produced inside the recurrence. Section~\ref{sec:better-ptq-algorithm} derives a conditional group cost and evaluates scales by recurrence-exact intra-group replay; Section~\ref{sec:infra-row-parallel} exploits output-row independence across tensor-parallel ranks.

% 给正向图画反向图 计算填空 为什么反向（选择） 为什么构建计算图（问答）
\section{Method}
\label{sec:method}

Our approach consists of two complementary components. The first is \emph{Schur Replay}, which improves the PTQ objective and group-scale selection algorithm. The second is a separate scalable execution infrastructure that preserves the numerical accuracy of the underlying quantizer.

\subsection{Better PTQ Algorithm}
\label{sec:better-ptq-algorithm}

We introduce \emph{Schur Replay}, a group-scale selection method for microscaled PTQ. Its central distinction from independent block reconstruction is that each candidate scale is evaluated under the conditional quadratic metric induced by the committed prefix and the compensation still available in the unquantized future. A closed-form update locates a promising region of the hardware-valid scale grid, after which recurrence-exact replay inside the current group selects the scale that is actually committed.

\subsubsection{NVFP4 Group Quantization}
\label{sec:schur-setup}

We use the notation and GPTQ recurrence of Section~\ref{sec:preliminaries}, with Hessia matrix $H=XX^\top$ and $H^{-1}=U^\top U$. Because Equation~\ref{eq:prelim-gptq-recurrence} is sequential across input columns, quantizing column $i$ changes the values subsequently presented to columns $i+1,\ldots,N$; scale selection must therefore account for the state induced inside a shared-scale group.

For NVFP4, each group $G$ contains $B=16$ consecutive input-column indices and shares a positive scale $s$. For a single output row, let $w_0\in\mathbb{R}^{B}$ denote the compensated group values upon entry. With $\rho:\mathbb{R}\to\Lambda$ denoting elementwise round-to-nearest onto the E2M1 levels $\Lambda=\{0,\pm0.5,\pm1,\pm1.5,\pm2,\pm3,\pm4,\pm6\}$, the group quantizer is
\begin{equation}
    Q_s(w)=s\,\rho(w/s).
    \label{eq:nvfp4-quantizer}
\end{equation}
The effective multiplier $s$ must be representable by the E4M3 block scale, and the largest reconstructed magnitude is $6s$. If $\gamma>0$ converts a stored positive E4M3 code to its effective multiplier, then
\begin{equation}
    \mathcal{S}=\{a/\gamma:a\in\mathrm{E4M3}^{+}\}.
    \label{eq:exportable-grid}
\end{equation}
Searching directly on $\mathcal{S}$ ensures that deployment-time scale alignment cannot alter the selected solution.

\subsubsection{Conditional Group Metric}
\label{sec:schur-metric}

At the entrance to a group, partition the GPTQ-ordered column indices into the committed prefix weight $P$, current group weight $G$ of 16 columns, and unquantized future weight $F$, and write $\mathcal{R}=G\cup F$ for the trailing indices. The prefix is fixed, and its optimal affine compensation has already been propagated into $w_0$. For candidate scale $s$, recurrence replay (defined below) produces $q(s)\in\mathbb{R}^{B}$; define the \emph{actual} group quantization error as $\delta_G(s)=q(s)-w_0$. The future weight remains free to compensate this error; writing $\delta_F\in\mathbb{R}^{|F|}$ for an arbitrary adjustment on the future coordinates, the candidate's marginal reconstruction cost is
\begin{equation}
\begin{aligned}
    \Delta f_G(s)
    &=\min_{\delta_F}
    \begin{bmatrix}\delta_G(s)\\\delta_F\end{bmatrix}^{\!\top}
    \begin{bmatrix}H_{GG}&H_{GF}\\H_{FG}&H_{FF}\end{bmatrix}
    \begin{bmatrix}\delta_G(s)\\\delta_F\end{bmatrix} \\
    &=\delta_G(s)^\top S_{G\mid P}\,\delta_G(s),\\
    S_{G\mid P}&=H_{GG}-H_{GF}H_{FF}^{-1}H_{FG}.
    \label{eq:schur-loss}
\end{aligned}
\end{equation}
Here $Z_{AB}$ denotes a submatrix of $Z$, and we abbreviate $S_{G\mid P}$ as $S_G$. The displayed blocks are the trailing curvature after the committed prefix has been fixed and its affine correction absorbed into $w_0$. With $\Omega=H^{-1}=U^\top U$, removing the prefix from the inverse system and then eliminating the free future gives
\begin{equation}
 H_{\mathcal{R}\mathcal{R}}^{-1}=\Omega_{\mathcal{R}\mathcal{R}}-\Omega_{\mathcal{R}P}\Omega_{PP}^{-1}\Omega_{P\mathcal{R}}=U_{\mathcal{R}\mathcal{R}}^\top U_{\mathcal{R}\mathcal{R}};
 \qquad
 S_{G\mid P}^{-1}=(H_{\mathcal{R}\mathcal{R}}^{-1})_{GG}=U_{GG}^\top U_{GG}.
 \label{eq:schur-from-cholesky}
\end{equation}
The second identity eliminates the free future weight $F$. Equivalently, $S_{G\mid P}$ is the conditional curvature seen by $G$ if it is left until last among the trailing variables.

Appendix~\ref{app:schur-proof} proves both identities. The reason for using the Schur complement is that scale selection should measure the loss that remains after the still-unquantized coordinates respond optimally, rather than the isolated error of the current block. Directly scoring $\lVert\delta_G(s)\rVert_2^2$ or $\delta_G(s)^\top H_{GG}\delta_G(s)$ implicitly sets $\delta_F=0$ and therefore charges the candidate for error components that later GPTQ updates can compensate. Minimizing the trailing quadratic instead gives the optimal response $\delta_F^\star=-H_{FF}^{-1}H_{FG}\delta_G(s)$; substituting it back yields
\begin{equation}
    \Delta f_G(s)
    =\delta_G(s)^\top\!\left(H_{GG}-H_{GF}H_{FF}^{-1}H_{FG}\right)\!\delta_G(s)
    =\delta_G(s)^\top S_{G\mid P}\delta_G(s).
    \label{eq:schur-marginal-interpretation}
\end{equation}
Thus the Schur score is exactly the \emph{marginal reconstruction error} made irreversible after finished quantization of current block, after the quantized prefix is fixed and the future is allowed its best continuous compensation. It accounts jointly for the realized 16-column error and conditional cross-column curvature. This interpretation also clarifies the surrogate boundary: the future is optimized continuously when candidates are ranked, whereas later blocks will eventually be quantized. We therefore use the score to compare candidate commitments at the same GPTQ state, not as a claim that it equals the final discrete layer loss. For one column, $U_{GG}=[u_{ii}]$ and $S_{G\mid P}=1/u_{ii}^2$, so the score reduces to $\delta_i^2/u_{ii}^2=e_i^2$, the scalar OBQ/GPTQ marginal cost \citep{frantar2022obc,frantar2023gptq}. The Schur quadratic is therefore its block generalization under the same future-compensation principle.

\subsubsection{Scale Proposal by Conditional Quadratic Minimization}
\label{sec:scale-proposal}

The discrete E2M1 codes depend on $s$, but if a code vector $Q\in\Lambda^B$ is held fixed, minimizing the block marginal cost in Equation~\ref{eq:schur-loss} along the scale coordinate is a one-dimensional quadratic problem:
\begin{equation}
    \phi(s)=(w_0-sQ)^\top S_G(w_0-sQ),
    \qquad
    \boxed{s^*(Q)=\frac{Q^\top S_Gw_0}{Q^\top S_GQ}}.
    \label{eq:closed-form-scale}
\end{equation}
This update applies the same quadratic-minimization principle underlying OBS-family weight compensation \citep{hassibi1993obs,frantar2022obc}, but along a shared scale coordinate rather than an individual weight coordinate. Because $Q=\rho(w_0/s)$, we resolve the scale--code dependence with $J=3$ fixed-point iterations initialized by the scale $s_{\mathrm{old}}$ currently assigned to the group,
\begin{equation}
    s^{(t+1)}
    =\frac{\rho(w_0/s^{(t)})^\top S_Gw_0}
    {\rho(w_0/s^{(t)})^\top S_G\rho(w_0/s^{(t)})},
    \qquad s^{(0)}=s_{\mathrm{old}},\quad t=0,\ldots,J-1.
    \label{eq:scale-fixed-point}
\end{equation}
We retain $s^{(t)}$ when the denominator is numerically zero and clamp a valid update to positive values. This fixed-code step is deliberately a proposal rather than the final decision. Its quadratic update is inexpensive and curvature-aware, but changing $s$ can change the E2M1 codes, and GPTQ compensation can then change the values rounded later in the group. Committing the proposal directly would therefore optimize an internally inconsistent snapshot. We instead use it only to center a small neighborhood on the exportable grid; the next stage evaluates that neighborhood under the actual sequential recurrence.

\subsubsection{Recurrence-Exact Replay on the Exportable Grid}
\label{sec:schur-replay}

To evaluate a scale without the fixed-code approximation, we replay the actual GPTQ recurrence inside $G$. Let $d_j=(U_{GG})_{jj}$, let $w_k^{(j)}$ denote the working value at group coordinate $k$ after replaying columns $1,\ldots,j$, and initialize $w^{(0)}=w_0$. For candidate $s$,
\begin{equation}
\begin{aligned}
    q_j(s)&=s\,\rho\!\left(w_j^{(j-1)}/s\right),
    &e_j(s)&=\frac{w_j^{(j-1)}-q_j(s)}{d_j},\\
    w_k^{(j)}&=w_k^{(j-1)}-e_j(s)(U_{GG})_{jk},
    &&k=j+1,\ldots,B.
    \label{eq:group-replay}
\end{aligned}
\end{equation}
The vector $q(s)=(q_1(s),\ldots,q_B(s))^\top$ collects the replayed quantized values. Its actual error, sign-reversed residual, and conditional marginal score are
\begin{equation}
    \delta_G(s)=q(s)-w_0,
    \qquad r(s)=-\delta_G(s),
    \qquad L(s)=\delta_G(s)^\top S_G\delta_G(s)=r(s)^\top S_Gr(s).
    \label{eq:replay-loss}
\end{equation}
Every candidate starts from the same block-entry state $w_0$ and uses the commit-time E2M1 rounding implementation. This common initialization is essential: otherwise candidates evaluated later would inherit corrections produced by earlier candidates and would no longer be comparable. Replay maintains a private working state for each candidate, reproduces all intra-group dependencies, and returns both the codes that would be committed and their realized error. Equation~\ref{eq:schur-loss} then compares those realized errors under the continuous-future marginal surrogate; only the winning candidate updates the global GPTQ state.

\begin{figure*}[t]
    \centering
    \includegraphics[width=0.98\textwidth]{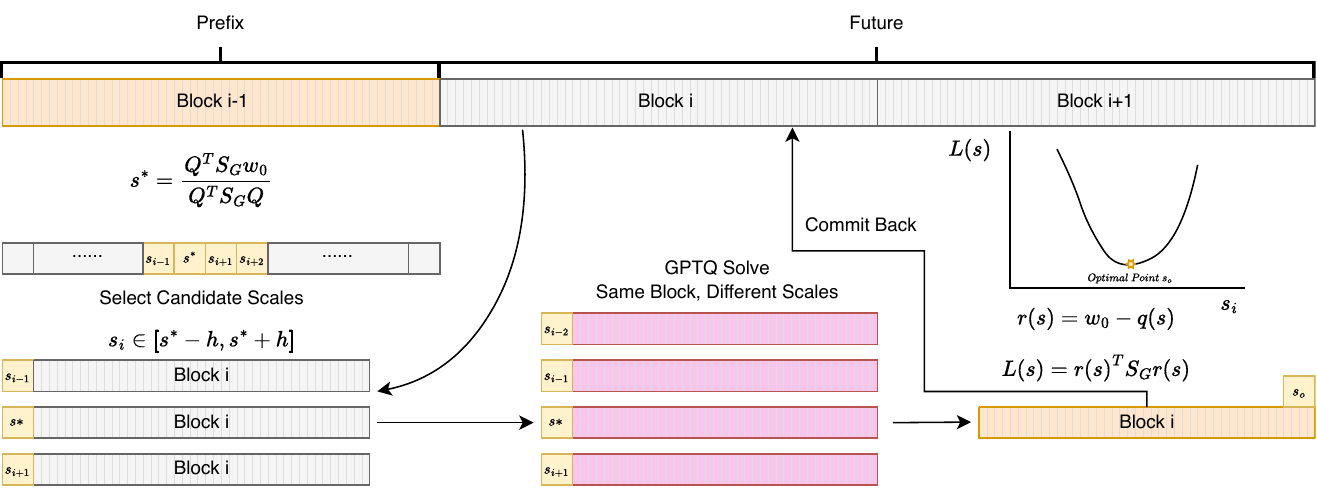}
    \caption{Schur Replay proposes a local window of exportable scales, replays each candidate from the same block-entry state, and commits the candidate minimizing $L(s)=\delta_G(s)^\top S_{G\mid P}\delta_G(s)$.}
    \label{fig:schur-replay-overview}
\end{figure*}

Exhaustively replaying the full E4M3 grid is unnecessary. Order its $C=|\mathcal S|$ admissible scales as $\mathcal{S}=\{\sigma_1<\cdots<\sigma_C\}$, let $s^*=s^{(J)}$ be the final proposal from Equation~\ref{eq:scale-fixed-point}, and define the clipped insertion index $c=\operatorname{clip}(\operatorname{searchsorted}(\mathcal{S},s^*))\in\{1,\ldots,C\}$. For a chosen nonnegative window half-width $h$, we evaluate
\begin{equation}
    \mathcal{C}
    =\{\sigma_{\operatorname{clip}(c+o)}:o=-h,\ldots,h\}
    \cup\{\Pi_{\mathcal{S}}(s_{\mathrm{old}})\},
    \qquad
    \widehat{s}=\operatorname*{arg\,min}_{s\in\mathcal{C}}L(s),
    \label{eq:candidate-window}
\end{equation}
where $\operatorname{clip}$ restricts indices to $\{1,\ldots,C\}$ and $\Pi_{\mathcal S}$ projects to the nearest exportable scale. The fixed-code update locates a promising neighborhood but does not decide the final scale: recurrence-exact replay adjudicates among the hardware-valid candidates using the state each candidate actually creates. Figure~\ref{fig:schur-replay-overview} summarizes this proposal--replay--commit sequence.

\paragraph{Exhaustive-window audit.}
An exhaustive audit over all 126 exportable scales finds that the $h=4$ window contains a global surrogate optimum for 83.28\% of Qwen rows and 83.53\% of Llama rows; $h=8$ exceeds 99.96\% for both. Appendix~\ref{app:window-audit} reports layer-wise results and the sampled-group scope.

Retaining the status-quo scale gives the per-group surrogate guarantee
\begin{equation}
    L(\widehat{s})\le L\!\left(\Pi_{\mathcal{S}}(s_{\mathrm{old}})\right).
    \label{eq:monotonicity}
\end{equation}
Ties within tolerance prefer the old scale and then the smaller scale, making selection deterministic. Because the projected old scale is always included in $\mathcal C$, the local search cannot increase the conditional replay surrogate relative to retaining that scale, even when the fixed-point proposal lands in a poor neighborhood. This is a per-group guarantee for the calibration-defined surrogate rather than a monotonicity claim for downstream task accuracy. Setting $h\ge C$ recovers the exhaustive-grid optimum.

Algorithm~\ref{alg:schur-replay} gives the full procedure. For $K\le 2h+2$ candidates and fixed $B=16$, proposal and replay add $O(JB^2+KB^2)$ work beyond the $O(B^3)$ group metric. Replay is vectorized over output rows and candidates as described in Section~\ref{sec:infra-parallel-commit}. We use $h=4$ for both model families, capping each group at ten candidates without model- or dataset-specific tuning.

\subsection{Faster PTQ Framework}
\label{sec:faster-ptq-infra}

\begin{figure*}[t]
    \centering
    \includegraphics[width=0.99\textwidth]{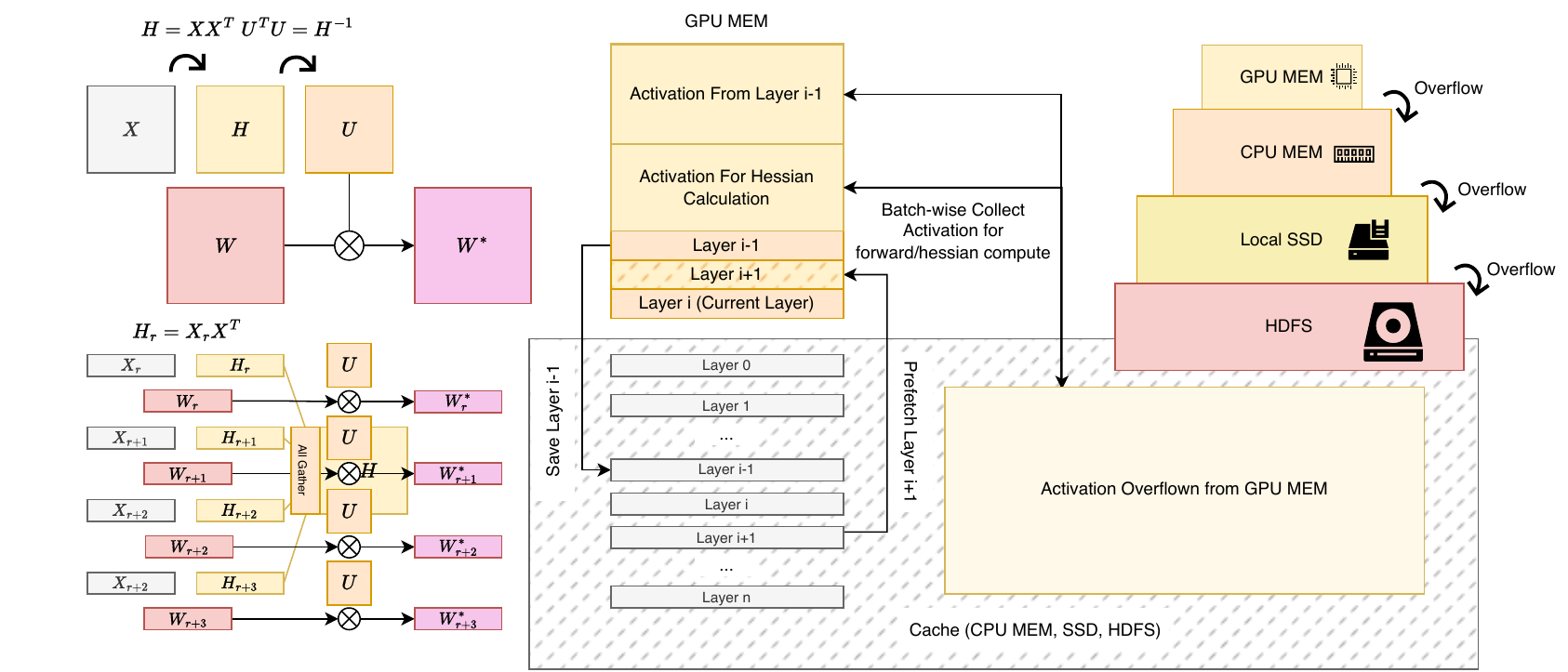}
    \caption{Scalable NVFP4 PTQ infrastructure: shared-Hessian row-parallel reconstruction (left), layer-wise activation collection and weight prefetch/retirement (center), and GPU--CPU--SSD--HDFS overflow (right).}
    \label{fig:ptq-infrastructure-overview}
\end{figure*}

Applying reconstruction-based PTQ to large dense and mixture-of-experts models creates a systems problem: weights, activations, and second-order state can exhaust memory, while assigning complete layers to devices leaves a large-layer solve serial. Our infrastructure combines \emph{layer-wise memory virtualization} with \emph{same-layer output-row-parallel GPTQ}. The former manages a complete per-layer lifecycle across storage tiers; the latter distributes independent row recurrences across tensor-parallel ranks. Both preserve the mathematical row-wise solve: they change placement and scheduling, not the quantization objective or update order within a row (Figure~\ref{fig:ptq-infrastructure-overview}).

\subsubsection{Batched Candidate Replay and Parallel Commit}
\label{sec:infra-parallel-commit}

Output rows and candidate scales are independent despite the column-sequential GPTQ recurrence. We therefore solve a tensor of shape $M\times K\times B$: each row--candidate pair owns a private trailing state, while one batched launch processes all pairs at a fixed column. The implementation still executes the $B$ columns in their original order, so batching exposes independent work without replacing the recurrence by an independent-block approximation; it requires $B$ sequential steps rather than $KB$ separately launched steps.

After scale selection, the commit path quantizes only the current group and propagates only the winning error to the unquantized suffix. The trailing correction is an in-place rank-one update, and CUDA Graph capture reduces launch overhead without changing the eager kernel sequence. Algorithm~\ref{alg:batched-replay-commit} gives the schedule.

\subsubsection{Aggressive Layer-Wise Memory Virtualization}
\label{sec:infra-tiered-storage}

GPU memory prioritizes the current layer, activation Gram state, and inverse-Hessian factor; CPU memory, rank-local SSD, and finally HDFS hold lower-priority activation and reconstruction state. This ordering keeps solver-critical tensors resident while permitting calibration activations to exceed device and host capacity. Temporary spill files are atomically published so interrupted writes cannot be consumed as valid state.

As shown in Figure~\ref{fig:ptq-infrastructure-overview}, only layer $i$ is installed on device during calibration; a pinned two-slot pool prefetches layer $i+1$, and the full-precision layer is retired after its NVFP4 fragment becomes durable. Consequently, device weight residency is bounded by the active layer plus one prefetched layer rather than the full model, leaving memory for Hessian factorization and replay. All the prefetch and offload of activations and weights are executed asynchornously while the main solver executes on the current layer, Algorithm~\ref{alg:tiered-residency} and Appendix~\ref{app:infra-details} give the protocol.

\subsubsection{Recoverable Streaming Export}
\label{sec:infra-recovery}

Each calibrated layer is packed immediately into a local NVFP4 fragment. As the quantized previous layer are always immediately stored in the storage when quantization is finished, we can use these cache as kind of a checkpoint. 

An atomic manifest advances only after the fragment and next-layer inputs are durable, enabling layer-grained recovery and full-precision retirement. Appendix~\ref{app:infra-details} and Algorithm~\ref{alg:recoverable-layerwise-ptq} give the details.

\subsubsection{Same-Layer Output-Row-Parallel GPTQ}
\label{sec:infra-row-parallel}

As shown in the left of Figure~\ref{fig:ptq-infrastructure-overview}, for a row-parallel layer, rank $r$ holds $W_r\in\mathbb{R}^{M\times N_r}$ and $X_r\in\mathbb{R}^{N_r\times T}$. After all-gathering $X=\operatorname{cat}_0(X_0,\ldots,X_{P_{\mathrm{tp}}-1})$, each rank forms $H_r=X_rX^\top$; gathering these row blocks yields the shared $H=XX^\top$. Because the reconstruction objective separates over output rows, ranks can solve disjoint row ranges with the same activation order and inverse-Hessian factor, gather the solved rows, and re-shard along $N$. Thus all ranks work on the current layer rather than assigning that entire layer to one owner. This decomposition equals a single-device solve in exact arithmetic; Appendix~\ref{app:infra-details} reports floating-point validation.

Column-parallel layers reduce Hessian numerators and sample counts before renormalization; grouped experts retain separate Hessians and quantizer states. Algorithm~\ref{alg:row-parallel-gptq} gives the schedule.

\section{Experiments}
\label{sec:experiments}

\subsection{Setup and Baselines}
\label{sec:exp-setup}

\paragraph{Models, calibration, and evaluation.}
We evaluate W4A4 NVFP4 on Qwen3.5-397B-A17B \citep{qwen2026qwen35} and Llama-3.3-70B-Instruct \citep{meta2024llama33,dubey2024llama3}. Accuracy runs share a decontaminated 255-document science/commonsense/math JSONL (seed 42), deterministic load order, batch size 1, and maximum length 16,384; Appendices~\ref{app:calibration-protocol} and~\ref{app:quantized-modules} specify the model checkpoints, calibration construction, quantization hyperparameters, module coverage, and structural exceptions. The 397B model uses EP8/TP8 and the 70B model uses one GPU for quantization and TP2 for inference. We report single-sample greedy pass@1 ($T=0$, top-$p=1$, and at most 8,192 new tokens) over 15,461 questions from MMLU-Pro, GSM8K, CMath, LiveCodeBench, English MGSM, HumanEval, and GPQA-Diamond \citep{wang2024mmlupro,cobbe2021gsm8k,wei2023cmath,jain2024livecodebench,shi2022mgsm,chen2021humaneval,rein2023gpqa}; Appendix~\ref{app:benchmark-protocol} gives dataset versions, prompts, grader provenance, exact counts, and paired uncertainty.
\begin{equation}
    R_{\mathrm{w}} = 100\,
    \frac{\sum_{d} n_d s^{\mathrm{quant}}_d}
         {\sum_{d} n_d s^{\mathrm{BF16}}_d},
    \label{eq:weighted-recovery}
\end{equation}
where $n_d$ is the number of questions in dataset $d$.  Values slightly above $100\%$ indicate higher aggregate accuracy than BF16 on this finite evaluation set.

\paragraph{Baselines.}
We compare GPTQ with SmoothQuant and AWQ rescaling; Four Over Six, SOAR, and ScaleSweep block-scale selection; ARCQuant residual channels; MR-GPTQ micro-rotations; and Schur Replay \citep{xiao2023smoothquant,lin2024awq,cook2026fouroversix,bao2026soar,lin2026scalesweep,meng2026arcquant,egiazarian2026mrgptq}. SmoothQuant, AWQ, Four Over Six, SOAR, and ScaleSweep use the same GPTQ reconstruction backend; ARCQuant and MR-GPTQ follow their format-specific reconstruction procedures.

\subsection{W4A4 Accuracy}
\label{sec:exp-accuracy}

Tables~\ref{tab:qwen-w4a4} and~\ref{tab:llama-w4a4} report all seven task scores. Schur Replay has the highest quantized question-weighted recovery: $99.35\%$ on Qwen3.5-397B-A17B and $100.84\%$ on Llama-3.3-70B-Instruct, versus ScaleSweep's $97.28\%$/$99.03\%$ and MR-GPTQ's $93.86\%$/$91.81\%$.

\begin{table}[H]
    \centering
    \caption{W4A4 results on Qwen3.5-397B-A17B.  \emph{W. Rec.} is the question-weighted recovery in Equation~\ref{eq:weighted-recovery}. Bold marks the best quantized method; BF16 provides the full-precision reference.}
    \label{tab:qwen-w4a4}
    \setlength{\tabcolsep}{2.2pt}
    \renewcommand{\arraystretch}{0.80}
    \fontsize{8}{8.5}\selectfont
    % Auto-generated by figures/generate_experiment_tables.py; do not edit.
\begin{tabular}{@{}lcccccccc@{}}
\toprule
Method & MMLU-P & GSM8K & CMath & LiveCode & MGSM & HumEval & GPQA & W. Rec. (\%) \\
\midrule
BF16 & 76.09 & 92.19 & 91.80 & 44.75 & 92.00 & 91.46 & 55.05 & 100.00 \\
\midrule
MR-GPTQ & 70.99 & 89.54 & 87.89 & 40.75 & 88.40 & 91.46 & 43.43 & 93.86 \\
GPTQ & 73.12 & 91.36 & 92.90 & 41.75 & 91.20 & 85.37 & 45.96 & 96.69 \\
ScaleSweep & 73.60 & 92.12 & 91.89 & 41.75 & 90.00 & 93.29 & 47.98 & 97.28 \\
Four Over Six+GPTQ & 73.91 & 90.67 & 91.53 & 40.25 & 89.20 & 87.80 & 52.02 & 97.32 \\
SmoothQuant+GPTQ & 74.07 & 90.22 & 91.53 & 43.00 & 89.20 & 92.68 & 52.53 & 97.60 \\
AWQ+GPTQ & 74.09 & 91.13 & 90.71 & 41.50 & 92.80 & 93.29 & 53.03 & 97.69 \\
ARCQuant & 75.17 & 89.99 & 89.80 & 39.75 & 90.80 & 90.24 & 48.48 & 98.34 \\
SOAR+GPTQ & 74.55 & 92.04 & 92.81 & 41.25 & 89.20 & 92.68 & 55.05 & 98.38 \\
\textbf{Schur Replay (ours)} & 75.40 & 93.03 & 92.17 & 42.25 & 92.80 & 92.68 & 52.53 & \textbf{99.35} \\
\bottomrule
\end{tabular}

\end{table}

\begin{table}[H]
    \centering
    \caption{W4A4 results on Llama-3.3-70B-Instruct under the same seven-dataset protocol as Table~\ref{tab:qwen-w4a4}.}
    \label{tab:llama-w4a4}
    \setlength{\tabcolsep}{2.2pt}
    \renewcommand{\arraystretch}{0.80}
    \fontsize{8}{8.5}\selectfont
    % Auto-generated by figures/generate_experiment_tables.py; do not edit.
\begin{tabular}{@{}lcccccccc@{}}
\toprule
Method & MMLU-P & GSM8K & CMath & LiveCode & MGSM & HumEval & GPQA & W. Rec. (\%) \\
\midrule
BF16 & 55.19 & 93.25 & 86.61 & 31.00 & 92.00 & 84.15 & 37.88 & 100.00 \\
\midrule
MR-GPTQ & 49.33 & 89.76 & 85.70 & 29.75 & 91.20 & 84.76 & 36.87 & 91.81 \\
GPTQ & 54.37 & 91.81 & 85.97 & 29.25 & 91.20 & 84.15 & 42.42 & 98.67 \\
AWQ+GPTQ & 54.24 & 93.18 & 86.70 & 32.00 & 92.80 & 81.71 & 39.90 & 98.85 \\
ScaleSweep & 54.53 & 92.72 & 86.34 & 31.50 & 90.00 & 84.15 & 38.38 & 99.03 \\
Four Over Six+GPTQ & 54.90 & 91.51 & 84.70 & 29.25 & 90.00 & 82.32 & 44.44 & 99.14 \\
SmoothQuant+GPTQ & 54.75 & 91.89 & 86.98 & 30.75 & 92.40 & 82.93 & 37.37 & 99.26 \\
ARCQuant & 55.44 & 93.56 & 83.70 & 32.50 & 93.60 & 82.32 & 40.40 & 100.15 \\
SOAR+GPTQ & 55.68 & 92.19 & 85.79 & 30.00 & 91.20 & 81.71 & 34.34 & 100.20 \\
\textbf{Schur Replay (ours)} & 55.93 & 93.18 & 86.16 & 29.25 & 92.00 & 83.54 & 39.90 & \textbf{100.84} \\
\bottomrule
\end{tabular}

\end{table}

\subsection{Ablations}
\label{sec:exp-ablations}

With ten candidates ($h=4$) for both models, Qwen macro recovery at $h=1,4,8,16$ is $97.70\%$, $98.92\%$, $98.62\%$, and $97.21\%$. Although $h=8$ doubles the neighborhood and raises surrogate-optimum coverage from $83.3\%$ to $99.96\%$, it does not improve recovery. Removing Schur scoring, recurrence replay, or both loses 90, 15, or 115 answers; static scoring raises held-out log-probability MSE from 1.769 to 1.842 ($t=3.2$). Appendix~\ref{app:ablation-tables} gives the full W4A4 ablations and W4A16 comparison.

\subsection{Infrastructure Efficiency}
\label{sec:exp-infra}

All 397B GPTQ pipelines share a checkpoint; ours and ModelOpt use comparable Megatron TP8/EP8 representations, while LLM Compressor uses its Hugging Face path. Only ours parallelizes the current layer's solve across ranks. With identical $32\times4096$ inputs, steady-layer times are 120.25, 1,824.29, and 2,782.89 seconds for ours, ModelOpt, and LLM Compressor; the 70B control gives $1.13\times$/$2.76\times$ speedups. Measured 70B GPTQ totals are 0.80, 0.91, and 2.09 hours, while the corresponding 397B totals (2.01, 30.63, and 46.42 hours) are projections; peak memory is 35.0, 218.6, and 86.6\,GB/GPU. For reference, we additionally report calibration time for our Schur Replay algorithm: 54.91 seconds per steady layer and 1.22 hours end-to-end for 70B on one GPU, and 157.84 seconds per steady layer and 2.67 hours end-to-end for 397B on eight GPUs. Because Schur Replay performs additional per-group scale search, these measurements are not included in the infrastructure speedup ratios. Appendix~\ref{app:infra-details} gives checks.

\begin{figure}[t]
    \centering
    \includegraphics[width=0.99\textwidth]{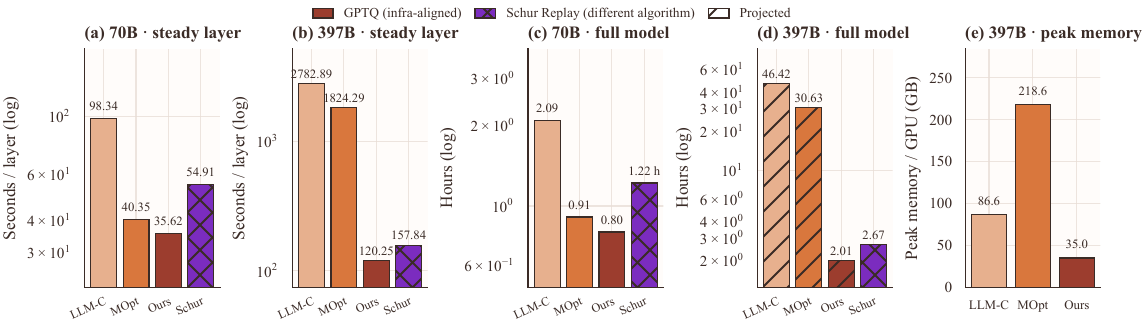}
    \caption{NVFP4 quantization efficiency. Panels (a--b) report steady-layer compute, (c--d) full-model compute, and (e) 397B peak memory. Warm bars compare infrastructure under aligned GPTQ; diagonal hatching marks projected 397B GPTQ totals. Purple cross-hatched bars are separate measured Schur Replay runs: W4A16 on one GPU for 70B and on eight GPUs for 397B. The 397B GPTQ bars use W4A4, so the purple bars are contextual measurements rather than inputs to the infrastructure speedup ratios.}
    \label{fig:infra-compute}
\end{figure}

\FloatBarrier

\section{Conclusion}
\label{sec:conclusion}

Schur Replay provides recurrence-exact scale evaluation and conditional Schur scoring; our separate execution infrastructure provides memory-bounded, output-row-parallel GPTQ. Together, these components produce 397B and 70B NVFP4 models that retain $99.35\%$ and $100.84\%$ question-weighted BF16 recovery across seven benchmarks, with ablations supporting both replay and scoring. The infrastructure reduces measured per-layer and 70B full-model quantization compute against the evaluated public toolchains, and the 397B run peaks at 35.0\,GB per GPU; 397B full-model runtimes remain layer-type-aware projections. These results show that hardware-valid NVFP4 conversion can be accurate and scalable without retraining.

The current evidence is empirical rather than universal: the exhaustive scale audit covers the sampled groups, and downstream evaluation spans two architectures and seven tasks. Future work should test whether the same reconstruction gains transfer to other microscaling formats, calibration regimes, and model families, while extending complete end-to-end measurements to the largest deployments.

\clearpage
\subsection*{AI use statement}
Generative AI tools were used to assist with code drafting, debugging, and language polishing of the manuscript. All AI-assisted code and text were reviewed and verified by the authors, who take full responsibility for the final implementation, experiments, claims, and manuscript content.

\subsection*{Ethics statement}
This work's controlled quantitative experiments use publicly available models and public benchmarks and do not involve human subjects or private data. We additionally mention an anonymized, authorized integration in an internal production pipeline, but do not disclose or release the proprietary model, data, metric definitions, or raw outputs, and do not use that integration to support the paper's quantitative claims. Reducing the cost of deploying large language models may broaden access to their capabilities. However, quantization does not remove biases, safety risks, or other limitations inherited from the underlying models, and deployments should retain the safeguards appropriate to those models and their application contexts.

\subsection*{Reproducibility statement}
The main paper specifies the quantization objective, scale-selection procedure, calibration protocol, evaluation metric, and infrastructure comparisons. The appendix provides complete pseudocode, derivations, model and module coverage, calibration details, ablation definitions, exhaustive-window audits, and additional system measurements. The source package contains the data-derived tables and figures used in the manuscript.

\clearpage
\bibliographystyle{plainnat}
\bibliography{tex/ref}

\clearpage
\appendix
\section{Code \& Artifact Availability}
\label{app:code-availability}

All development in this work is built on NVIDIA ModelOpt. Upon acceptance, we will release a general-purpose patch that can be installed on top of ModelOpt to enable Schur Replay for both W4A4 and W4A16 quantization. The patch will also provide the core infrastructure contributions introduced in this paper, allowing ModelOpt to execute PTQ methods---including GPTQ, GPTAQ, and Schur Replay---with improved hardware efficiency.

We will additionally release the calibration dataset used for the public-model experiments, together with its deterministic construction metadata, and all quantized model artifacts produced for the main experiments, subject to the licenses and redistribution terms of the corresponding base models and source datasets. These artifacts are intended to support direct reproduction of the reported public-model evaluation without requiring users to repeat every large-scale quantization run.

The released implementation will support the complete four-tier GPU--CPU--SSD--HDFS storage hierarchy used in our public-model experiments when both a local SSD path and an HDFS path are configured. If either persistent-storage path is unavailable, the same implementation will automatically degrade to a three-tier cache hierarchy while preserving the quantization algorithm and execution semantics. The proprietary Prompt Enhancer, internal evaluation assets, and production-specific integration are outside the release scope.

\clearpage
\section{Complete Ablation Tables}
\label{app:ablation-tables}

\begin{table}[ht]
    \centering
    \caption{Replay-window ablation on Qwen3.5-397B-A17B W4A4.}
    \label{tab:window-ablation}
    \small
    % Auto-generated; do not edit.
\begin{tabular}{lrr}
\toprule
Half-width $h$ & Default & Macro recovery (\%) \\
\midrule
1 & no & 97.70 \\
\textbf{4} & yes & \textbf{98.92} \\
8 & no & 98.62 \\
16 & no & 97.21 \\
\bottomrule
\end{tabular}

\end{table}

\paragraph{Factorial design and counting unit.}
We cross two residual constructions with two scoring metrics while holding the candidate grid, scale proposal, old-scale retention, tie-breaking, calibration data, and quantization backend fixed. The \emph{static} construction independently rounds every column from the shared group-entry snapshot $w_0$,
\begin{equation}
    q_j^{\mathrm{static}}(s)=s\,\rho(w_{0,j}/s),
    \qquad r^{\mathrm{static}}(s)=w_0-q^{\mathrm{static}}(s).
    \label{eq:static-ablation-residual}
\end{equation}
Although the implementation receives $U_{GG}$ through the common loss interface, static construction does not use it: the quantization error of column $j$ never changes the value rounded at column $j+1$. The \emph{replay} construction instead executes the commit-time GPTQ recurrence in Equation~\ref{eq:group-replay}. It rounds the updated state $w_j^{(j-1)}$ and propagates each normalized error through $(U_{GG})_{j,j+1:B}$ before quantizing subsequent columns, producing
\begin{equation}
    r^{\mathrm{replay}}(s)=w_0-q^{\mathrm{replay}}(s).
    \label{eq:replay-ablation-residual}
\end{equation}
The residual axis therefore tests entry-snapshot RTN codes against the codes realized by the sequential GPTQ commit path. Independently, the metric axis scores either residual with the conditional Schur loss $r^\top S_Gr$ or the Euclidean loss $\lVert r\rVert_2^2$. The four cells are \emph{replay $\times$ Schur} (ours), \emph{static $\times$ Schur} (\texttt{score\_rtn}), \emph{replay $\times$ Euclidean} (\texttt{metric\_id}), and \emph{static $\times$ Euclidean} (\texttt{pure\_rtn}). We additionally report BF16 and replay $\times$ Schur with $h=16$ (\texttt{win16}). Because dataset percentages have different resolution, Table~\ref{tab:factorial-counts} reports absolute correct-question counts; ``1q'' is the percentage-point change caused by one question within that dataset.

All four cells are internal ablations of our own pipeline and none reproduces a published method. Every cell retains the exportable grid of Equation~\ref{eq:exportable-grid}, the closed-form proposal of Equation~\ref{eq:closed-form-scale}, the same local candidate window, and old-scale retention; only the scored quantity changes. In particular \texttt{metric\_id} is not a prior baseline but our search with the conditional metric removed, and \texttt{score\_rtn} applies the Schur metric to a residual the commit path never produces. Baseline comparisons against published methods appear in Tables~\ref{tab:qwen-w4a4} and~\ref{tab:llama-w4a4}.

Attributing the metric main effect requires distinguishing two uses of second-order information. Established GPTQ/OBC-style reconstruction uses $H^{-1}$ to order columns and to normalize per-column error during weight compensation. The metric axis here instead evaluates a shared block scale under $S_G$, the Schur complement that eliminates the still-unquantized future columns; Equation~\ref{eq:schur-from-cholesky} shows this equals $(U_{GG}^\top U_{GG})^{-1}$ and reduces to the classical scalar score only at $B=1$. This extension of the conditional metric from single-weight decisions to shared-scale selection is introduced in this work, so the dominant metric effect measures the contribution of that extension rather than of preexisting practice.

\begin{table}[ht]
    \centering
    \caption{Per-dataset absolute correct counts for the controlled metric $\times$ residual-construction ablation. The two panels contain the same rows and together specify every cell. Bold marks the best quantized count per dataset.}
    \label{tab:factorial-counts}
    \setlength{\tabcolsep}{3.0pt}
    \renewcommand{\arraystretch}{0.93}
    \begin{tabular}{@{}lrrrrr@{}}
        \toprule
        Dataset & $n_q$ & Weight & 1q (pp) & BF16 & Ours \\
        \midrule
        MMLU-Pro & 12,032 & 77.8\% & 0.008 & 9,155 & 9,072 \\
        GSM8K & 1,319 & 8.5\% & 0.076 & 1,216 & \textbf{1,227} \\
        CMath & 1,098 & 7.1\% & 0.091 & 1,008 & 1,012 \\
        LiveCodeBench & 400 & 2.6\% & 0.250 & 179 & \textbf{169} \\
        MGSM-English & 250 & 1.6\% & 0.400 & 230 & \textbf{232} \\
        HumanEval & 164 & 1.1\% & 0.610 & 150 & 152 \\
        GPQA-Diamond & 198 & 1.3\% & 0.505 & 109 & 104 \\
        \midrule
        Total & 15,461 & 100\% & -- & 12,047 & 11,968 \\
        \bottomrule
    \end{tabular}

    \vspace{0.35em}
    \begin{tabular}{@{}lrrrr@{}}
        \toprule
        Dataset & \texttt{score\_rtn} & \texttt{metric\_id} & \texttt{pure\_rtn} & \texttt{win16} \\
        \midrule
        MMLU-Pro & 9,079 & 9,010 & 8,979 & \textbf{9,093} \\
        GSM8K & 1,218 & 1,206 & 1,212 & 1,216 \\
        CMath & 1,006 & \textbf{1,013} & 1,011 & \textbf{1,013} \\
        LiveCodeBench & 161 & 166 & 166 & 167 \\
        MGSM-English & 225 & 228 & 229 & 223 \\
        HumanEval & \textbf{157} & 149 & 154 & 148 \\
        GPQA-Diamond & \textbf{107} & 106 & 102 & 100 \\
        \midrule
        Total & 11,953 & 11,878 & 11,853 & 11,960 \\
        \bottomrule
    \end{tabular}
\end{table}

\begin{table}[ht]
    \centering
    \caption{Question-weighted recovery and deficit relative to the 12,047-correct BF16 reference.}
    \label{tab:factorial-summary}
    \setlength{\tabcolsep}{4pt}
    \renewcommand{\arraystretch}{0.94}
    \begin{tabular}{@{}lrrr@{}}
        \toprule
        Cell & Correct & Recovery (\%) & $\Delta$ questions vs. BF16 \\
        \midrule
        BF16 & 12,047 & -- & -- \\
        Ours (replay $\times$ Schur) & 11,968 & 99.344 & $-79$ \\
        \texttt{score\_rtn} (static $\times$ Schur) & 11,953 & 99.220 & $-94$ \\
        \texttt{win16} (replay $\times$ Schur, $h=16$) & 11,960 & 99.278 & $-87$ \\
        \texttt{metric\_id} (replay $\times$ Euclidean) & 11,878 & 98.597 & $-169$ \\
        \texttt{pure\_rtn} (static $\times$ Euclidean) & 11,853 & 98.390 & $-194$ \\
        \bottomrule
    \end{tabular}
\end{table}

\paragraph{Main effects and interaction.}
Starting from ours, replacing Schur with Euclidean scoring while holding replay fixed loses 90 questions ($-0.747$ recovery points). Replacing recurrence replay with a static residual while holding Schur fixed loses 15 questions ($-0.124$ points). An additive model therefore predicts a joint loss of 105 questions, or 11,863 correct and $-0.871$ points. The measured static $\times$ Euclidean cell has 11,853 correct (98.390\% recovery), ten fewer questions than this prediction. Thus both Schur scoring and recurrence replay contribute positively in the controlled aggregate, the metric main effect is larger, and their joint removal has an additional $-10$-question ($-0.083$-point) interaction.

\subsection{Weight-Only NVFP4 (W4A16)}
\label{app:w4a16}

The main results quantize both weights and activations. To separate the weight-reconstruction contribution from activation quantization, Table~\ref{tab:w4a16} repeats the Qwen3.5-397B-A17B comparison with weights in NVFP4 and activations in BF16, holding the calibration set, evaluation suite, greedy decoding protocol, and BF16 reference fixed. Because the archived per-question scores cover the same seven non-floor datasets and 15,461 questions as Tables~\ref{tab:qwen-w4a4} and~\ref{tab:llama-w4a4}, both recovery aggregates are recomputed here on that scope rather than reusing the 10-dataset macro recorded in the source file.

\begin{table}[ht]
    \centering
    \caption{Weight-only NVFP4 (W4A16) on Qwen3.5-397B-A17B over the same seven datasets and 15,461 questions as the W4A4 tables. ``Rel.\ MSE'' is the relative weight-reconstruction error; ``Macro'' and ``Weighted'' are unweighted and question-weighted recovery against the 12,047-correct BF16 reference.}
    \label{tab:w4a16}
    \small
    % Auto-generated by figures/generate_w4a16_table.py; do not edit.
\begin{tabular}{@{}llcrrr@{}}
\toprule
Method & Act-order & Rel. MSE & Correct & Macro (\%) & Weighted (\%) \\
\midrule
BF16 reference & -- & -- & 12,047 & 100.00 & 100.00 \\
\midrule
RTN + shrink search & -- & 0.0832 & 11,656 & 94.65 & 96.75 \\
GPTQ & -- & 0.0480 & 11,908 & 95.48 & 98.84 \\
GPTQ & \checkmark & 0.0440 & 11,599 & 94.99 & 96.28 \\
Schur Replay & -- & 0.0379 & 11,997 & 98.49 & 99.59 \\
\textbf{Schur Replay} & \textbf{\checkmark} & \textbf{0.0365} & \textbf{12,055} & \textbf{99.18} & \textbf{100.07} \\
\bottomrule
\end{tabular}

\end{table}

Two observations carry over from W4A4. Schur Replay attains the lowest weight-reconstruction error and the highest recovery in both act-order settings, and its advantage over GPTQ is larger than the gap between the two GPTQ variants. The ordering of reconstruction error also matches the ordering of recovery across all five arms here, which the W4A4 setting does not always exhibit because activation quantization adds an error source that weight reconstruction cannot address. Act-order interacts differently with the two methods: it helps Schur Replay on both metrics but lowers GPTQ recovery from $98.84\%$ to $96.28\%$ despite improving its reconstruction error, so reconstruction error alone does not determine end-to-end accuracy. The single arm exceeding $100\%$ weighted recovery reflects small gains on GSM8K and HumanEval offsetting losses on LiveCodeBench and GPQA-Diamond, and single-run greedy decoding does not resolve differences of this size; the per-dataset counts and uncertainty treatment in Appendix~\ref{app:benchmark-protocol} apply here as well.

\section{Pseudocode}
\label{app:algorithms}

\newcounter{algorithm}
\renewcommand{\thealgorithm}{\arabic{algorithm}}
\newcommand{\algorithmcaption}[1]{%
  \refstepcounter{algorithm}%
  \par\smallskip\noindent\textbf{Algorithm~\thealgorithm:} #1\par
}

This appendix collects the complete pseudocode for Schur Replay and the execution infrastructure described in Section~\ref{sec:method}. We use $\operatorname{AG}$ for all-gather, $\operatorname{cat}_d$ for concatenation along dimension $d$, $[n]=\{0,\ldots,n-1\}$, and $\epsilon>0$ for a numerical guard.

\begin{figure}[H]
\small
\centering
\begin{tabular}{p{0.95\linewidth}}
\textbf{Schur Replay group-scale selection}\\
\hline
$\mathbf{Input:}\;w_0\!\in\!\mathbb{R}^{M\times B},\ U_{GG}\!\in\!\mathbb{R}^{B\times B},\ s_0\!\in\!\mathcal{S}^{M},\ h,\ J=3$\\
$S_G\leftarrow(U_{GG}^{\top}U_{GG})^{-1};\quad s\leftarrow s_0$\\
\textbf{for} $t=1,\ldots,J$ \textbf{do}\\
\hspace{1em}$Q\leftarrow\rho(w_0/s);\quad d\leftarrow\operatorname{diag}(QS_GQ^\top)$\\
\hspace{1em}$s\leftarrow(QS_Gw_0^\top)_{\mathrm{diag}}/\max(d,\epsilon)$ if $d>\epsilon$\\
\textbf{end for}\\
$c_m\leftarrow\operatorname{clip}(\operatorname{searchsorted}(\mathcal{S},s_m));\quad \mathcal{C}_m\leftarrow\{\mathcal{S}_{\operatorname{clip}(c_m+o)}\}_{o=-h}^{h}\cup\{\Pi_{\mathcal{S}}(s_{0,m})\}$\\
$K\leftarrow2h+2;\quad \mathbf{C}\leftarrow\operatorname{pad}(\{\mathcal{C}_m\}_{m=1}^{M},K);\quad R\leftarrow w_0[:,\emptyset,:]\otimes\mathbf{1}_{K}$ \hfill \textit{// mask padded slots}\\
\textbf{for} $j=1,\ldots,B$ \textbf{do}\\
\hspace{1em}$q_j\leftarrow \mathbf{C}\,\rho(R_{:,:,j}/\mathbf{C});\quad e_j\leftarrow(R_{:,:,j}-q_j)/(U_{GG})_{jj}$\\
\hspace{1em}$R_{:,:,j+1:B}\leftarrow R_{:,:,j+1:B}-e_j\otimes(U_{GG})_{j,j+1:B}$\\
\textbf{end for}\\
$q\leftarrow\operatorname{stack}_{j=1}^{B}(q_j)$\\
$L_{m,k}\leftarrow(w_{0,m,:}-q_{m,k,:})S_G(w_{0,m,:}-q_{m,k,:})^\top$ for valid $\mathbf{C}_{m,k}$; padded losses are $+\infty$\\
$L_m(s)\leftarrow(w_{0,m,:}-q_m(s))S_G(w_{0,m,:}-q_m(s))^\top$\\
$k_m^*\leftarrow\arg\min_kL_{m,k};\quad \widehat{s}_m\leftarrow \mathbf{C}_{m,k_m^*};\quad \operatorname{commit}(w_0,\widehat{s},U_{GG})$\\
$\mathbf{assert}\;L_m(\widehat{s}_m)\le L_m(\Pi_{\mathcal{S}}(s_{0,m})),\ \forall m$\\
\hline
\end{tabular}
\algorithmcaption{The fixed-code update proposes a local window, and recurrence-exact group replay selects and commits one exportable NVFP4 scale per output-row group.}
\label{alg:schur-replay}
\end{figure}

\begin{figure}[H]
\small
\centering
\begin{tabular}{p{0.95\linewidth}}
\textbf{Batched replay and group-local GPTQ commit}\\
\hline
$\mathbf{Input:}\;W\!\in\!\mathbb{R}^{M\times N},\ U\!\in\!\mathbb{R}^{N\times N},\ \mathcal{G}=\{g,\ldots,g+B-1\},\ S_G\!\in\!\mathbb{R}^{B\times B},\ S\!\in\!\mathbb{R}^{M\times K}$\\
$W^{\mathrm{in}}\leftarrow W;\quad W_0\leftarrow W_{:,\mathcal{G}};\quad U_G\leftarrow U_{\mathcal{G},\mathcal{G}};\quad R\leftarrow W_0[:,\emptyset,:]\otimes\mathbf{1}_K$\\
\textbf{for} $j=1,\ldots,B$ \textbf{do}\\
\hspace{1em}$D_j\leftarrow\max((U_G)_{j,j},\epsilon);\quad \widehat{W}_{:,:,j}\leftarrow Q_{\mathrm{NVFP4}}(R_{:,:,j};S)$\\
\hspace{1em}$E_j\leftarrow(R_{:,:,j}-\widehat{W}_{:,:,j})/D_j$\\
\hspace{1em}$R_{:,:,j+1:B}\leftarrow R_{:,:,j+1:B}-E_j\otimes(U_G)_{j,j+1:B}$\\
\textbf{end for}\\
$\mathcal{L}_{m,k}\leftarrow(\widehat{W}_{m,k,:}-W_{0,m,:})S_G(\widehat{W}_{m,k,:}-W_{0,m,:})^\top$\\
$k_m^*\leftarrow\arg\min_{k\in\{1,\ldots,K\}}\mathcal{L}_{m,k};\quad S_m^*\leftarrow S_{m,k_m^*},\quad m=1,\ldots,M$\\
\textbf{for} $j=1,\ldots,B$ \textbf{do}; $i\leftarrow g+j-1$\\
\hspace{1em}$\widehat{G}_{:,j}\leftarrow Q_{\mathrm{NVFP4}}(W_{:,i};S^*);\quad e_i\leftarrow(W_{:,i}-\widehat{G}_{:,j})/U_{i,i}$\\
\hspace{1em}$W_{:,i:N}\leftarrow W_{:,i:N}-e_i\otimes U_{i,i:N}$\\
\textbf{end for}\\
$\mathbf{Invariant:}\; (\widehat{G},W_{:,g+B:N})\equiv_{\mathrm{bit}}\operatorname{Commit}_{\mathrm{eager}}(W^{\mathrm{in}},\mathcal{G},S^*;U)$\\
\hline
\end{tabular}
\algorithmcaption{Candidate-scale batching exposes the independent row and candidate dimensions without changing the column-wise GPTQ recurrence.}
\label{alg:batched-replay-commit}
\end{figure}

\begin{figure}[H]
\small
\centering
\begin{tabular}{p{0.95\linewidth}}
\textbf{Tiered capture and asynchronous layer residency}\\
\hline
$\mathbf{Input:}\; \{x_b\}_{b=0}^{N_b-1}\ \text{(activation batches)},\ K_{\mathrm{gpu}},\ K_{\mathrm{host}}\ \text{(tier capacities)},\ \{\theta_i\}_{i=0}^{L-1}$\\
$\tau(b)\leftarrow
\begin{cases}
\mathrm{GPU}, & b<K_{\mathrm{gpu}},\\
\mathrm{CPU}, & K_{\mathrm{gpu}}\le b<K_{\mathrm{gpu}}+K_{\mathrm{host}},\\
\mathrm{DISK},& \text{otherwise};
\end{cases}\quad C_b\leftarrow\operatorname{put}_{\tau(b)}(x_b)$\\
$\mathcal{H}\leftarrow\operatorname{park}_{\mathrm{CPU}}(\{\theta_i\});\quad \mathcal{A}\leftarrow\emptyset$\\
\textbf{Prefetch}$(i)$: $u\leftarrow i\bmod2$; $P_u\leftarrow\operatorname{pin}(\mathcal{H}_i)$; $\widetilde{\theta}_i\leftarrow\operatorname{H2D}_{\mathrm{async}}(P_u)$; record event $\eta_i$\\
\textbf{Onload}$(i)$: $\operatorname{wait}(\eta_i)$; $\theta_i\leftarrow\widetilde{\theta}_i$; $\mathcal{A}\leftarrow\mathcal{A}\cup\{i\}$\\
\textbf{Release}$(i)$: $\theta_i\leftarrow\operatorname{offload}(\theta_i)$ or $\emptyset$ after $\operatorname{pack}(i)$; $\mathcal{A}\leftarrow\mathcal{A}\setminus\{i\}$\\
$\mathbf{Invariant:}\;|\mathcal{A}|\le1,\quad \widetilde{\theta}_{i+1}\notin\mathcal{A},\quad \operatorname{consume}(C_b)\Rightarrow C_b\notin\mathrm{DISK}$\\
\hline
\end{tabular}
\algorithmcaption{Bounded activation storage and single-layer weight residency prevent full-model weights and the complete calibration cache from residing on the accelerator simultaneously.}
\label{alg:tiered-residency}
\end{figure}

\begin{figure}[H]
\small
\centering
\begin{tabular}{p{0.95\linewidth}}
\textbf{Recoverable layer-wise PTQ pipeline}\\
\hline
$\mathbf{Input:}\;\{\ell_i=(\theta_i,\cdot)\}_{i=0}^{L-1},\ C_0\ \text{(input cache)},\ w\ \text{(commit window)}$\\
$\mathcal{M}\ \text{(manifest)};\quad i_0\leftarrow\operatorname{resume}(\mathcal{M});\quad \mathcal{F}_r\leftarrow\emptyset$ on every rank $r$\\
$\operatorname{park}(\{\ell_i\});\quad \operatorname{prefetch}(i_0)$\\
\textbf{for} $i=i_0,\ldots,L-1$ \textbf{do}\\
\hspace{1em} $(z_{<i-1},z_{i-1},z_i)\leftarrow(\textsc{Skip},\textsc{Run},\textsc{Capture})$\\
\hspace{1em} $X_i\leftarrow\operatorname{materialize}(C_i)$; $\operatorname{onload}(i)$; $\operatorname{prefetch}(i+1)$\\
\hspace{1em} $(a_i,H_i)\leftarrow\operatorname{calibrate}(\ell_i,X_i)$; $(\widehat{\theta}_i,q_i,m_i)\leftarrow\operatorname{PTQ}(\theta_i;a_i,H_i)$\\
\hspace{1em} $(F_i,h_i)\leftarrow\operatorname{atomic\_pack}(\widehat{\theta}_i)$; $C_{i+1}\leftarrow\operatorname{capture\_next}(\ell_i,\widehat{\theta}_i,C_i)$\\
\hspace{1em}$\mathcal{F}_r\leftarrow\mathcal{F}_r\cup\{F_i\}$; $\operatorname{save}(\widehat{\theta}_i,q_i,m_i)$\\
\hspace{1em} \textbf{if} $(i+1)\bmod w=0$ or $i=L-1$: $\operatorname{commit}(\mathcal{M},i,C_{i+1})$\\
\hspace{1em} $\operatorname{discard}(\theta_i)$\\
\textbf{end for}\\
$\min_r|\mathcal{F}_r|=\max_r|\mathcal{F}_r|=L;\quad A\leftarrow\operatorname{finalize}(\{(F_i,h_i)\}_{i=0}^{L-1})$\\
$\mathbf{Invariant:}\;\mathcal{M}.i=\max\{j:\{0{:}j\}\ \text{is durable}\},\quad \operatorname{SHA256}(F_i)=h_i$\\
\hline
\end{tabular}
\algorithmcaption{Atomic window commits and hash-validated streaming fragments make long-running PTQ recoverable without retaining the full-precision model in memory.}
\label{alg:recoverable-layerwise-ptq}
\end{figure}

\begin{figure}[H]
\small
\centering
\begin{tabular}{p{0.95\linewidth}}
\textbf{Row-parallel GPTQ solve}\\
\hline
$\mathbf{Input:}\;N_r=N/P_{\mathrm{tp}},\ B\mid N_r,\ \mathcal{J}_r=[rN_r,(r+1)N_r),\ \mathcal{B}_r=[rN_r/B,(r+1)N_r/B),\ r\in[P_{\mathrm{tp}}]$\\
$\lambda\ge0\ \text{(absolute Hessian damping)};\quad B=16$\\
$W_r\!\in\!\mathbb{R}^{M\times N_r},\quad A_r\!\in\!\mathbb{R}^{M\times N_r/B}\ \text{(block-scale metadata)},\quad X_r\!\in\!\mathbb{R}^{N_r\times T}$\\
$X\leftarrow\operatorname{cat}_{0}(\operatorname{AG}(X_r))\in\mathbb{R}^{N\times T}$ \hfill \textit{// gather contraction features; retain tokens}\\
$H_r\leftarrow X_rX^\top\in\mathbb{R}^{N_r\times N}$ \hfill \textit{// local Hessian row block}\\
$\{H_p\}_{p=0}^{P_{\mathrm{tp}}-1}\leftarrow\operatorname{AG}(H_r)$ \hfill \textit{// all-gather row blocks to every rank}\\
$H\leftarrow\operatorname{cat}_{0}(H_0,\ldots,H_{P_{\mathrm{tp}}-1})=XX^\top\in\mathbb{R}^{N\times N}$\\
$W\leftarrow\operatorname{cat}_{1}(\operatorname{AG}(W_r));\quad A\leftarrow\operatorname{cat}_{1}(\operatorname{AG}(A_r))$\\
$\pi\leftarrow\operatorname{actorder}(H);\quad H^\pi\leftarrow H_{\pi,\pi}$\\
$U\leftarrow\operatorname{chol}^{\top}((H^\pi+\lambda I_N)^{-1})$\\
$\mathcal{I}_r\leftarrow\operatorname{row\_partition}([M],r,P_{\mathrm{tp}})$\\
$(\widehat{W}^{(r)},\widehat{A}^{(r)})\leftarrow\operatorname{GPTQ}(W_{\mathcal{I}_r,:},A_{\mathcal{I}_r,:};U,\pi)$\\
$\widehat{W}\leftarrow\operatorname{cat}_{0}(\operatorname{AG}(\widehat{W}^{(r)}));\quad \widehat{A}\leftarrow\operatorname{cat}_{0}(\operatorname{AG}(\widehat{A}^{(r)}))$\\
$W_r\leftarrow\widehat{W}_{:,\mathcal{J}_r};\quad A_r\leftarrow\widehat{A}_{:,\mathcal{B}_r}$\\
$\mathbf{Invariant:}$ every rank uses the same $(H^\pi,U,\pi)$; in exact arithmetic, gathering $(\widehat{W}^{(r)},\widehat{A}^{(r)})$ equals $\operatorname{GPTQ}(W,A;U,\pi)$\\
\hline
\end{tabular}
\algorithmcaption{The row-parallel schedule converts otherwise idle tensor-parallel ranks into independent GPTQ workers without approximating the objective.}
\label{alg:row-parallel-gptq}
\end{figure}

\section{Additional Analysis and Infrastructure Details}

\subsection{Partitioned Proof of the Conditional Metric}
\label{app:schur-proof}

All blocks below are taken after applying the column permutation used by GPTQ, so the processing order is $P,G,F$ and $\mathcal{R}=G\cup F$ contains the trailing indices. Let $\Omega=H^{-1}=U^\top U$ and partition the global upper-triangular factor as
\begin{equation}
U=\begin{bmatrix}U_{PP}&U_{P\mathcal{R}}\\0&U_{\mathcal{R}\mathcal{R}}\end{bmatrix}.
\end{equation}
Consequently,
\begin{equation}
\Omega_{PP}=U_{PP}^\top U_{PP},\quad
\Omega_{P\mathcal{R}}=U_{PP}^\top U_{P\mathcal{R}},\quad
\Omega_{\mathcal{R}\mathcal{R}}=U_{P\mathcal{R}}^\top U_{P\mathcal{R}}+U_{\mathcal{R}\mathcal{R}}^\top U_{\mathcal{R}\mathcal{R}}.
\end{equation}
Applying the block-inverse identity to $\Omega=H^{-1}$ gives the inverse of the corresponding principal block of $H$ as the Schur complement of $\Omega_{PP}$:
\begin{align}
H_{\mathcal{R}\mathcal{R}}^{-1}
&=\Omega_{\mathcal{R}\mathcal{R}}-\Omega_{\mathcal{R}P}\Omega_{PP}^{-1}\Omega_{P\mathcal{R}}\\
&=U_{P\mathcal{R}}^\top U_{P\mathcal{R}}+U_{\mathcal{R}\mathcal{R}}^\top U_{\mathcal{R}\mathcal{R}}
-U_{P\mathcal{R}}^\top U_{PP}(U_{PP}^\top U_{PP})^{-1}U_{PP}^\top U_{P\mathcal{R}}\\
&=U_{\mathcal{R}\mathcal{R}}^\top U_{\mathcal{R}\mathcal{R}}.
\end{align}
The last equality uses nonsingularity of the damped factor $U_{PP}$, for which $U_{PP}(U_{PP}^\top U_{PP})^{-1}U_{PP}^\top=I$. Thus the contribution from rows in $P$ is present in $\Omega_{\mathcal{R}\mathcal{R}}$ but is exactly removed by the Schur term; no commutation between principal submatrices and inversion is assumed.

Next partition the trailing curvature $H_{\mathcal{R}\mathcal{R}}$ by $\mathcal{R}=G\cup F$. Eliminating the still-free future $F$ gives the prefix-conditional group curvature
\begin{equation}
S_{G\mid P}=H_{GG}-H_{GF}H_{FF}^{-1}H_{FG},
\end{equation}
whose block-inverse identity is
\begin{equation}
(H_{\mathcal{R}\mathcal{R}}^{-1})_{GG}=S_{G\mid P}^{-1}.
\end{equation}
Because $U_{\mathcal{R}\mathcal{R}}$ remains upper triangular,
\begin{equation}
U_{\mathcal{R}\mathcal{R}}=\begin{bmatrix}U_{GG}&U_{GF}\\0&U_{FF}\end{bmatrix},
\qquad
(U_{\mathcal{R}\mathcal{R}}^\top U_{\mathcal{R}\mathcal{R}})_{GG}=U_{GG}^\top U_{GG}.
\end{equation}
Combining the last two displays proves $S_{G\mid P}^{-1}=U_{GG}^\top U_{GG}$ and hence Equation~\ref{eq:schur-from-cholesky}. The block $U_{GG}$ is extracted from the globally permuted factor after preceding elimination.

\subsection{Calibration and Quantizer Configuration}
\label{app:calibration-protocol}

\paragraph{Accuracy experiments.}
Qwen3.5-397B-A17B and Llama-3.3-70B-Instruct use one identical calibration set. It contains 255 multi-turn documents: 85 each from science (SciQ), commonsense (CommonsenseQA and PIQA), and mathematics (NuminaMath-CoT). The production mixture is built with seed 42; the loader then applies a deterministic seed-0 shuffle. Documents target 11,000 characters (measured median 11,327), are rendered with \texttt{apply\_chat\_template} without an added generation prompt, and include assistant answers. Tokenization uses batch size 1, truncation/padding at 16,384 tokens, and therefore produces 255 calibration batches per model.

Candidate records are filtered against the evaluation data using normalized 13-gram overlap over prompts and gold answers. The scan covers the registered evaluation suites, and the builder fails closed if the blocklist, domain balance, chat structure, renderability, token budget, or length-parity checks fail. This guards against calibration--evaluation contamination while keeping the two model families on the same source records. Both accuracy recipes use identity pre-transforms, NVFP4 groups of 16 weights, and the same calibration set; Qwen uses TP8/EP8 and Llama uses TP1 during quantization.

\paragraph{Scale and Hessian settings.}
For tensor-wide scale $\gamma$, let $a_{\max}=\max_{m,n}|W_{mn}|$. The implementation maps $a_{\max}$ to the largest product of a positive E4M3 scale code (448) and the E2M1 maximum (6):
\begin{equation}
\gamma=\frac{448\times 6}{a_{\max}},
\qquad
\mathcal S=\{a/\gamma:a\in\mathrm{E4M3}^{+}\}.
\end{equation}
The fixed-code proposal runs three iterations from the incoming scale. If $Q^\top S_GQ\le 10^{-12}$, the update retains the incoming scale; valid updates are clamped positive. Both headline arms use the same fixed replay half-window $h=4$, an engineering operating point between replay cost and local reconstruction fidelity; it is not tuned separately by model or dataset. The incoming exportable scale is always retained, and deterministic ties prefer that scale and then the smaller candidate. Hessians are accumulated in FP32, symmetrized, and damped at 1\% of the mean diagonal for the production Qwen recipe; failed Cholesky factorizations trigger increasing damping rather than an identity-Hessian substitution. The active GPTQ column permutation is applied before extracting the blocks used in Equation~\ref{eq:schur-from-cholesky}.

\paragraph{Systems timing calibration.}
All three 397B timing pipelines load the same compatible Hugging Face checkpoint. Ours converts it to Megatron TP8/EP8 grouped-expert modules before quantization, and ModelOpt performs an analogous conversion within its framework, so ModelOpt is the same-representation baseline; LLM Compressor provides no comparable conversion and runs its Hugging Face-native path. Neither baseline distributes one layer's GPTQ solve across tensor-parallel ranks, so their Megatron-side support does not shorten that layer's critical path. The steady-state comparison uses 32 byte-identical pretokenized sequences at length 4,096 in all three frameworks; using 255 sequences at length 16,384 made the baseline Hessian passes infeasible within the study budget. The measured 70B full-model comparison uses the complete 255-by-16,384 protocol. Pretokenized artifacts record model, calibration path, sample count, maximum length, batch size, shuffle seed, batch/token counts, tokenizer class, and vocabulary size.

\subsection{Quantized Module Coverage}
\label{app:quantized-modules}

\paragraph{Common scope.}
For both model families, W4A4 and W4A16 quantize the same weight-bearing modules with block-16 NVFP4 weights. W4A4 additionally quantizes their input activations to block-16 NVFP4, whereas W4A16 retains BF16 activations. Embeddings, the output language-model head, normalization layers, rotary-position components, MoE routing and gating modules, multi-token-prediction heads, and vision components remain in high precision or outside the evaluated language backbone. We do not otherwise exempt the first or last transformer block.

\paragraph{Qwen3.5-397B-A17B.}
Our module coverage follows the weight-selection policy of the official Qwen3.5-397B-A17B-FP8 release (\url{https://huggingface.co/Qwen/Qwen3.5-397B-A17B-FP8}). The 60-block hybrid backbone contains 45 Gated DeltaNet (GDN) blocks and 15 full-attention blocks, with full attention at zero-based block indices $3,7,\ldots,59$. In every block, we quantize both routed-expert projections and both shared-expert projections. In each GDN block, we quantize the fused input projection and output projection; in each full-attention block, we quantize the fused QKV projection and output projection. Under TP8/EP8, this corresponds to 360 selected modules per rank: 120 grouped routed-expert modules, 120 shared-expert projections, 90 GDN projections, and 30 full-attention projections. Two structural exceptions are important. First, the fused GDN input projection is segmented: its query, key, value, and gate rows are quantized, whereas the small beta and alpha state-gating rows remain in high precision to avoid perturbing the recurrent decay dynamics. Second, each routed expert is calibrated and quantized independently, with its own Hessian statistics and scales; scales are not shared across experts.

\paragraph{Llama-3.3-70B-Instruct.}
The 80-block dense backbone has four selected modules per block: the fused QKV projection, attention output projection, MLP up projection, and MLP down projection, for 320 dense modules in total. It has no GDN segmentation, routed experts, shared experts, or grouped-linear special case. Quantization uses TP1; TP2 is used only for inference evaluation.

\subsection{Benchmark Evaluation, Provenance, and Uncertainty}
\label{app:benchmark-protocol}

\paragraph{Decoding and prompts.}
Each model--dataset pair uses one deterministic greedy completion with temperature 0, top-$p=1$, and at most 8,192 new tokens; the recorded seed field is inactive under temperature-0 decoding. The seven sets contain 12,032 MMLU-Pro, 1,319 GSM8K, 1,098 CMath, 400 LiveCodeBench, 250 English MGSM, 164 HumanEval, and 198 GPQA-Diamond questions, totaling 15,461. Math prompts request step-by-step reasoning and a final answer in \texttt{\textbackslash boxed\{\}}; MMLU-Pro and deterministically shuffled GPQA options request only the option letter. HumanEval requests the full Python function, and LiveCodeBench requests a complete program. Every JSONL record stores \texttt{idx}, dataset, model, subset, sample, seed, full prediction, and gold. Across every headline arm and the Qwen window/metric/scoring ablations, \texttt{(idx,sample)} is unique, \texttt{sample=0}, and all IDs and gold fields agree with the collection-local BF16 record.

\paragraph{Graders and executable tasks.}
Scoring uses OpenCompass commit \texttt{63b0199ba6c7f251386c04b0215ea5c3016a0407}, whose GSM8K, MATH, and multiple-choice postprocessors are called by our scorer. HumanEval uses the 164-record \texttt{openai/openai\_humaneval} test split: its canonical solutions match our gold fields byte for byte. LiveCodeBench uses the 400-record \texttt{test.jsonl} at Hugging Face revision \texttt{0fe84c3912ea0c4d4a78037083943e8f0c4dd505}, covering 2023-05-07 through 2024-03-02. The executor extracts the longest fenced code block, recognizes Python or C++, compiles C++17 with \texttt{-O2}, and checks HumanEval unit tests or LiveCodeBench public and private stdin/stdout cases. Re-scoring BF16, SOAR, and Schur Replay exactly reproduces every reported dataset percentage for both models. A second complete execution of both code sets gives zero changes among 6,768 repeated per-question verdicts.

\begin{table}[t]
    \centering
    \caption{Exact correct-question counts for Qwen3.5-397B-A17B. SOAR is the strongest external quantized baseline by question-weighted recovery.}
    \label{tab:benchmark-counts}
    \setlength{\tabcolsep}{3.2pt}
    \renewcommand{\arraystretch}{0.92}
    \begin{tabular}{@{}lrrr@{}}
        \toprule
        Dataset & BF16 & SOAR & Schur Replay \\
        \midrule
        MMLU-Pro (12,032) & 9,155 & 8,970 & 9,072 \\
        GSM8K (1,319) & 1,216 & 1,214 & 1,227 \\
        CMath (1,098) & 1,008 & 1,019 & 1,012 \\
        LiveCodeBench (400) & 179 & 165 & 169 \\
        MGSM-English (250) & 230 & 223 & 232 \\
        HumanEval (164) & 150 & 152 & 152 \\
        GPQA-Diamond (198) & 109 & 109 & 104 \\
        \midrule
        Total (15,461) & 12,047 & 11,852 & 11,968 \\
        Accuracy (\%) & 77.919 & 76.657 & 77.408 \\
        Wilson 95\% CI (\%) & [77.258,78.565] & [75.984,77.318] & [76.742,78.060] \\
        \bottomrule
    \end{tabular}
\end{table}

\begin{table}[t]
    \centering
    \caption{Exact correct-question counts for Llama-3.3-70B-Instruct under the same protocol as Table~\ref{tab:benchmark-counts}.}
    \label{tab:benchmark-counts-llama}
    \setlength{\tabcolsep}{3.2pt}
    \renewcommand{\arraystretch}{0.92}
    \begin{tabular}{@{}lrrr@{}}
        \toprule
        Dataset & BF16 & SOAR & Schur Replay \\
        \midrule
        MMLU-Pro (12,032) & 6,641 & 6,699 & 6,730 \\
        GSM8K (1,319) & 1,230 & 1,216 & 1,229 \\
        CMath (1,098) & 951 & 942 & 946 \\
        LiveCodeBench (400) & 124 & 120 & 117 \\
        MGSM-English (250) & 230 & 228 & 230 \\
        HumanEval (164) & 138 & 135 & 137 \\
        GPQA-Diamond (198) & 75 & 68 & 79 \\
        \midrule
        Total (15,461) & 9,389 & 9,408 & 9,468 \\
        Accuracy (\%) & 60.727 & 60.850 & 61.238 \\
        Wilson 95\% CI (\%) & [59.955,61.494] & [60.078,61.616] & [60.467,62.003] \\
        \bottomrule
    \end{tabular}
\end{table}

\clearpage
\paragraph{Macro-average sensitivity.}
For completeness, Table~\ref{tab:macro-means} reports the unweighted arithmetic mean of the seven dataset accuracies alongside the question-weighted recovery used in the main text. The macro mean can be biased as a model-level summary in this suite: it assigns the same weight to MMLU-Pro's 12,032 questions and to HumanEval's 164 or GPQA-Diamond's 198, although the latter accuracy estimates have substantially larger sampling variance. Consequently, a few outcomes on a small dataset can move the macro ranking more than their share of the 15,461 evaluated questions. We therefore treat it as a complementary task-balanced view and retain question-weighted recovery as the headline aggregate.

\begin{table}[H]
    \centering
    \caption{Seven-dataset macro mean and question-weighted recovery (\%). Macro averages datasets uniformly; W. Rec. weights their question counts and normalizes by the corresponding BF16 total. Bold marks the best quantized method in each column.}
    \label{tab:macro-means}
    \setlength{\tabcolsep}{4.0pt}
    \renewcommand{\arraystretch}{0.94}
    % Auto-generated by figures/generate_experiment_tables.py; do not edit.
\begin{tabular}{@{}lrrrr@{}}
\toprule
Method & Qwen Macro & Qwen W. Rec. & Llama Macro & Llama W. Rec. \\
\midrule
BF16 & 77.62 & 100.00 & 68.58 & 100.00 \\
\midrule
GPTQ & 74.52 & 96.69 & 68.45 & 98.67 \\
SmoothQuant+GPTQ & 76.18 & 97.60 & 68.15 & 99.26 \\
AWQ+GPTQ & 76.65 & 97.69 & 68.65 & 98.85 \\
Four Over Six+GPTQ & 75.05 & 97.32 & 68.16 & 99.14 \\
SOAR+GPTQ & 76.80 & 98.38 & 67.27 & 100.20 \\
ARCQuant & 74.89 & 98.34 & \textbf{68.79} & 100.15 \\
ScaleSweep & 75.80 & 97.28 & 68.23 & 99.03 \\
MR-GPTQ & 73.21 & 93.86 & 66.77 & 91.81 \\
Schur Replay (ours) & \textbf{77.27} & \textbf{99.35} & 68.57 & \textbf{100.84} \\
\bottomrule
\end{tabular}

\end{table}

On Qwen3.5-397B-A17B, Schur Replay leads both quantized aggregates (77.27 macro; 99.35 weighted recovery). On the smaller Llama-3.3-70B-Instruct model, the macro scores are tightly clustered: ARCQuant reaches 68.79, AWQ+GPTQ 68.65, and Schur Replay 68.57, a 0.22-point span between the first and third methods, while Schur Replay leads question-weighted recovery at 100.84. This modest separation is consistent with a prior cross-scale quantization evaluation, which reports that accuracy and method rankings depend on model size, task, and bit width and shows several smaller-model settings with close aggregate scores across quantizers \citep{lee2024quantizedevaluation}. It is descriptive evidence rather than a monotone scaling claim: architecture and task mix also differ between our 70B dense and 397B MoE models.

\paragraph{Paired uncertainty.}
We align methods by dataset and question ID, and bootstrap the paired correctness difference with 100,000 resamples and fixed seed 20,260,907. On Qwen, Schur Replay exceeds SOAR by 0.750 percentage points; the 95\% paired-bootstrap interval is $[0.220,1.281]$ points, and the exact two-sided McNemar test over 786 SOAR-only versus 902 Schur-only successes gives $p=0.0051$. On Llama, Schur Replay exceeds SOAR by 0.388 points, with an 80\% paired-bootstrap interval of $[0.013,0.763]$ points. Recovery remains collection-local: exact counts give 99.344\%/100.841\% for Schur Replay on Qwen/Llama, consistent with the rounded headline values.
\clearpage
\subsection{Exhaustive Exportable-Scale Audit}
\label{app:window-audit}

We score the complete 126-point exportable grid for every sampled group and compare its row-wise global optimum with the proposal-centered window. The audit covers all 60 decoder layers of Qwen3.5-397B-A17B (MoE, TP8/EP8; 69,016 groups and 143.5M non-flat rows) and all 80 layers of Llama-3.3-70B-Instruct (dense, TP1; 5,324 groups and 85.5M non-flat rows). Let $d$ be the signed grid distance from proposal center $c$ to the nearest exhaustive optimum. The optimum lies at $c$ for 28.44\%/28.90\% of rows, within one step for 57.47\%/58.34\%, within four for 83.28\%/83.53\%, and within eight for 99.965\%/99.989\% on Qwen/Llama. At $h=4$, layer-wise coverage remains between 82.79--85.83\% on Qwen and 83.10--85.06\% on Llama, showing that the aggregate is not dominated by a few layers.

These coverage numbers do not order end-to-end accuracy. Widening from $h=4$ to $h=8$ raises coverage of the surrogate optimum from 83.28\% to 99.965\% and doubles the replayed neighborhood, while macro recovery changes from 98.92\% to 98.62\% in this sensitivity sweep. The two quantities measure different properties: coverage concerns the calibration-defined surrogate, whereas recovery is measured on downstream tasks. This audit characterizes the proposal and sensitivity around the fixed $h=4$ engineering operating point; neither coverage nor downstream benchmark outcomes were used as a post-hoc selection rule. The sweep does not establish a monotone relation between search width and downstream quality. Equation~\ref{eq:monotonicity} applies only to the per-group surrogate because the incoming scale remains available; it makes no corresponding guarantee for downstream accuracy.

\begin{figure*}[t]
    \centering
    \includegraphics[width=0.98\textwidth]{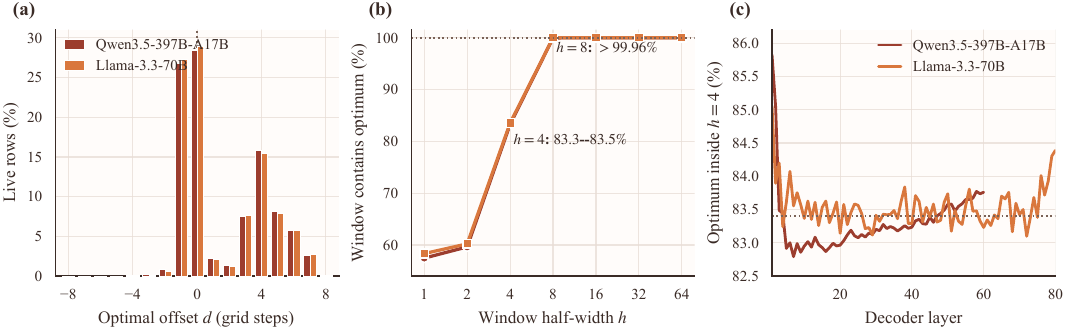}
    \caption{Exhaustive audit of the Schur Replay window over all 126 exportable scales. Percentages exclude flat rows for which every candidate ties. \textbf{(a)} Signed distance from the proposal center to the nearest optimum of the conditional replay surrogate. \textbf{(b)} Fraction of rows whose local window contains such an optimum. \textbf{(c)} Layer-wise coverage at $h=4$. The audit contains 143.5M live rows from Qwen3.5-397B-A17B and 85.5M from Llama-3.3-70B-Instruct.}
    \label{fig:schur-window-audit}
\end{figure*}

\subsection{Layer-Wise Recovery and Numerical Validation}
\label{app:infra-details}

\begin{figure*}[t]
    \centering
    \includegraphics[width=0.88\textwidth]{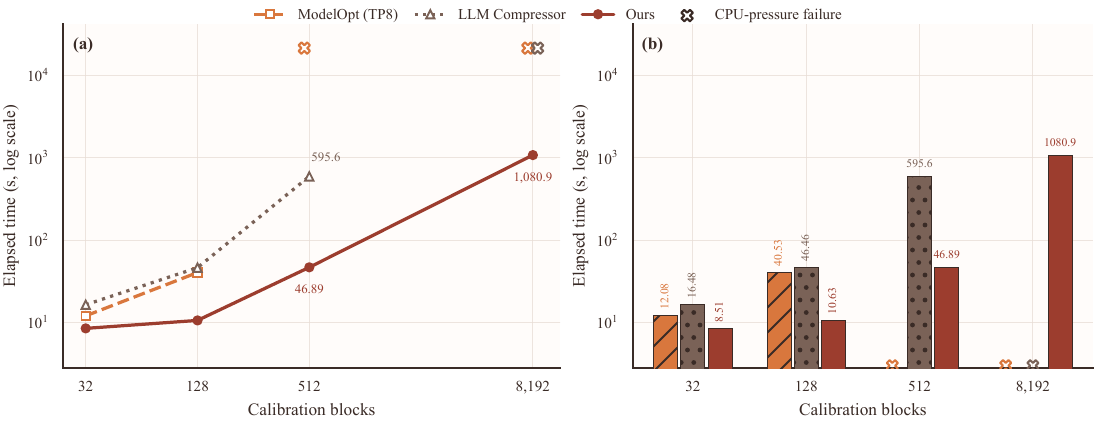}
    \caption{Scaling of a mock Qwen3.5 transformer layer with Qwen3.5-397B-A17B layer width under TP8 as calibration blocks increase at sequence length 16,384. The node has eight 72\,GB GPUs, 3.5\,TB of local SSD capacity, and mounted HDFS storage. \textbf{(a)} Scaling curves show completed elapsed-time measurements. \textbf{(b)} Grouped bars show the same measurements for direct comparison at each workload size. Both panels use logarithmic time axes; crosses mark runs that did not complete under host-memory pressure and are not assigned synthetic times or bar heights. Ours is faster for every jointly completed setting and is the only evaluated path that completes the 8,192-block stress test.}
    \label{fig:infra-mock-layer-scaling}
\end{figure*}

\paragraph{Calibration-state scaling stress test.}
Figure~\ref{fig:infra-mock-layer-scaling} evaluates activation and storage scaling on a mock Qwen3.5 transformer layer with the layer width of Qwen3.5-397B-A17B under TP8. The test uses one eight-GPU node with 72\,GB of device memory per GPU, 3.5\,TB of local SSD capacity, and a mounted HDFS-backed network volume. Sequence length is fixed at 16,384 while the number of calibration blocks increases from 32 to 8,192. The plotted times are the observed elapsed times of the reported runs; a cross denotes a run that did not complete under host-memory pressure and is not a timing value.

At 32 blocks, the optimized path takes 8.51 seconds, compared with 12.08 seconds for ModelOpt TP8 and 16.48 seconds for LLM Compressor. At 128 blocks, the corresponding times are 10.63, 40.53, and 46.46 seconds. At 512 blocks, ours completes in 46.89 seconds and LLM Compressor in 595.56 seconds, while ModelOpt fails under host-memory pressure. At 8,192 blocks, only ours completes, in 1,080.9 seconds; both public baselines fail under host-memory pressure. These measurements isolate a Qwen-width mock layer rather than an end-to-end model run, and support the infrastructure claim that tiered activation storage sustains calibration states beyond the host-memory envelope reached by the evaluated baselines.

The layer-wise forward uses three states. Layers older than the immediate predecessor become parameter-free skip modules that reconstruct zero-valued outputs from lightweight metadata; the preceding layer runs on cached inputs; and the current layer captures its inputs and terminates the model forward. This state machine preserves the active-layer residency bound even under framework hooks that would otherwise move parameters automatically. After packing, each retired parameter is replaced by a zero-element tensor rather than a meta tensor, so accidental reuse fails instead of propagating meta-valued outputs.

After each layer is calibrated, the system saves its weights, quantizer state, and output metadata. A window is committed only after the next-layer inputs are written and an atomic manifest records the last complete boundary. Resume validates the model depth and checkpoint interval to detect configuration drift. Each real-NVFP4 fragment and its source checkpoint, when present, carry SHA-256 digests; finalization memory-maps validated fragments, rejects duplicate module keys or conflicting absorbed tensors, builds the artifact locally, and verifies its digest after transfer. Distributed minimum and maximum reductions over completed-layer counts expose rank disagreement before finalization while retaining valid fragments for a resumed run.

All relocation, batching, graph replay, spill, checkpoint, and distributed-solve paths are checked against their reference implementations. Hessian inversion symmetrizes the matrix, handles structurally zero columns, and increases diagonal damping when Cholesky factorization fails; an unresolved failure invalidates the solve rather than substituting an identity Hessian. Because fused FP32 Hessian accumulation changes summation order, validation uses tight relative-error checks and requires identical sample counts and activation-order permutations.

\clearpage
\section{Sample Outputs}
\label{app:detailed-examples}
The examples below are organized by dataset. Within each benchmark, the Qwen3.5-397B-A17B disagreement is followed immediately by the Llama-3.3-70B-Instruct disagreement, and each model block compares SOAR, Four Over Six, and Schur Replay. We preserve archived generations verbatim apart from typography and explicitly marked omissions. When a Qwen generation reaches its length limit inside the reasoning trace, we label it incomplete rather than reconstructing a final answer.

\subsection{MMLU-Pro disagreements}

\paragraph{Qwen3.5-397B-A17B: MMLU-Pro, question 5798: reversing the good/bad percentages.}
The question asks what percentage of Germans described the relationship between the United States and Germany as good in 2019. The keyed option is A (34\%); option C gives 64\%. All three archived predictions contain closed reasoning traces.

\textbf{SOAR ($\times$), CoT excerpt.} ``The user is asking for a specific statistic from 2019 regarding German public opinion on the relationship between the U.S. and Germany. [\emph{source-identification discussion omitted}] A key finding from the 2019 Pew survey: `About two-thirds of Germans (64\%) say relations between the United States and Germany are good.' [\emph{checking omitted}] If 64\% said good, 34\% said bad. Option A is 34\%. This suggests the question is derived directly from this specific dataset where the distractors include the inverse statistic. [\emph{final recheck omitted}] Option A: 34\% (The \% who said it was bad). Option C: 64\% (The \% who said it was good). The question asks for `good'. So, C.'' \textbf{Emitted answer: C.}

\textbf{Four Over Six ($\times$), CoT excerpt.} ``The user is asking for a specific statistic from 2019 regarding German public opinion on the relationship between the U.S. and Germany. [\emph{source-identification discussion omitted}] In 2018, 79\% said it was good. In 2019, it dropped to 64\%. [\emph{checking omitted}] Pew Research Center, `Transatlantic Relations': 2017: 83\%; 2018: 79\%; 2019: 64\%. This seems consistent. Therefore, the answer is C. [\emph{option-by-option check omitted}] Conclusion: C is the correct answer.'' \textbf{Emitted answer: C.}

\textbf{Schur Replay ($\checkmark$), CoT excerpt.} ``The user is asking for a specific statistic from 2019 regarding German public opinion on US--Germany relations. [\emph{source-identification discussion omitted}] `Just 34\% of Germans describe relations between the U.S. and Germany as good.' This seems to be the specific data point. Therefore, Option A is the correct answer. [\emph{checking omitted}] Conclusion: 34\% is the most likely intended answer based on the prominent Pew Research data from that year. [\emph{option check omitted}] Good: 34\%. Bad: 64\%. Answer: A.'' \textbf{Emitted answer: A.}

\textbf{Observed disagreement.} The two baselines settle on the reversed 64\% statistic, whereas Schur Replay distinguishes the keyed good/bad percentages and emits A.

\paragraph{Llama-3.3-70B-Instruct.}
\medskip
\noindent\textbf{Question 122 (business).}
Which of the following is not one of the underlying principles of the corporate governance Combined Code of Practice?

\textbf{Options.} A: Integrity; B: Adaptability; C: Accountability; D: Openness; E: Dependability; F: Acceptability.

\begin{center}
\begin{tabular}{@{}lcccc@{}}
\toprule
& Gold & SOAR & Four Over Six & Schur Replay \\
\midrule
Emitted letter & F & E $\times$ & E $\times$ & F $\checkmark$ \\
\bottomrule
\end{tabular}
\end{center}

\medskip
\noindent\textbf{Question 6224 (health).}
A deficiency of which vitamin has been associated with enamel defects and increased risk of dental caries?

\textbf{Options.} A: Vitamin D; B: Vitamin K; C: Riboflavin; D: Niacin; E: Biotin; F: Vitamin A; G: Vitamin E; H: Vitamin C; I: Vitamin B12; J: Folic acid.

\begin{center}
\begin{tabular}{@{}lcccc@{}}
\toprule
& Gold & SOAR & Four Over Six & Schur Replay \\
\midrule
Emitted letter & A & F $\times$ & F $\times$ & A $\checkmark$ \\
\bottomrule
\end{tabular}
\end{center}

\medskip
\noindent\textbf{Question 9028 (physics).}
When the shadow of the Moon falls on Earth, we have a:

\textbf{Options.} A: lunar eclipse; B: lunar eclipse if daytime and solar eclipse if nighttime; C: very dangerous event; D: planetary alignment; E: celestial phenomenon; F: supernova event; G: solar eclipse if daytime and lunar eclipse if nighttime; H: solar flare event; I: solar eclipse; J: moonlight night.

\begin{center}
\begin{tabular}{@{}lcccc@{}}
\toprule
& Gold & SOAR & Four Over Six & Schur Replay \\
\midrule
Emitted letter & I & G $\times$ & G $\times$ & I $\checkmark$ \\
\bottomrule
\end{tabular}
\end{center}

\textbf{Observed difference.} Across three categories, both baselines converge to the same distractor while Schur Replay emits the keyed option. For these three particular Llama records, each displayed letter is the complete archived generation; no reasoning text was stored for these records. This does not imply that every Llama MMLU-Pro prediction is short, whereas the Qwen MMLU-Pro case above explicitly preserves its archived reasoning excerpts.

\subsection{GSM8K disagreements}

\paragraph{Qwen3.5-397B-A17B: GSM8K, question 1019: whether late checkouts have departed.}
A hotel starts with 100 guests. Twenty-four check out early and 15 choose late checkout. In the afternoon, twice as many people check in as chose late checkout, and seven more check in after dinner. The gold answer is 98.

\textbf{SOAR ($\times$), CoT excerpt.} ``Initial state: Hotel has 100 guests. [\emph{intervening deliberation omitted}] A `late checkout' implies that these guests are staying in their rooms longer than the standard checkout time. They have not left the hotel yet. [\emph{arithmetic omitted}] $100-24+30+7=113$.'' Its final response repeats this interpretation and emits $\boxed{113}$.

\textbf{Four Over Six ($\checkmark$), CoT excerpt.} ``Initial state: Hotel has 100 guests. [\emph{intervening deliberation omitted}] Since the timeline of the problem extends to `after dinner' (evening), guests with a late checkout (typically 1 PM--4 PM) would have left by this time. [\emph{arithmetic omitted}] Result: 98.'' Its final response computes $100-24-15+30+7$ and emits $\boxed{98}$.

\textbf{Schur Replay ($\checkmark$), CoT excerpt.} ``Initial state: Hotel has 100 guests. [\emph{analysis omitted}] 24 guests check out early. [\emph{step omitted}] 15 guests check out late. Since the question asks for the count `now,' after the sequence of events including the afternoon and dinner, these guests have departed.'' Its final response obtains $61+30+7$ and emits $\boxed{98}$.

\textbf{Observed disagreement.} SOAR treats the late-checkout group as still present at the final time and obtains 113; Four Over Six and Schur Replay account for their departure and recover the keyed 98.

\paragraph{Llama-3.3-70B-Instruct.}
\textbf{Question.} John buys 2 pairs of shoes for each of his 3 children. They cost \$60 each. How much did he pay?

\textbf{Gold derivation.} He bought $3\times2=6$ pairs of shoes, so he spent $60\times6=\$360$.

\textbf{SOAR ($\times$), complete generation.} To find the total amount John paid, we need to calculate the total number of shoes he bought and then multiply it by the cost of each shoe. (1) John buys 2 pairs of shoes for each of his 3 children. Since each pair contains 2 shoes, the total number of shoes per child is $2\times2=4$ shoes. (2) John has 3 children, so the total number of shoes he buys is $4\times3=12$ shoes. (3) Each shoe costs \$60, so the total amount John paid is $12\times\$60=\$720$. Therefore, the total amount John paid is $\boxed{720}$.

\textbf{Four Over Six ($\times$), complete generation.} To find the total amount John paid, we need to calculate the total number of shoes he bought and then multiply it by the cost of each shoe. Step 1: John buys 2 pairs of shoes for each of his 3 children. Since each pair contains 2 shoes, the total number of shoes per child is $2\text{ pairs}\times2\text{ shoes/pair}=4$ shoes. For 3 children, the total number is $4\text{ shoes/child}\times3\text{ children}=12$ shoes. Step 2: Each shoe costs \$60. Therefore, the total amount John paid is $12\text{ shoes}\times\$60/\text{shoe}=\$720$. The final answer is $\boxed{720}$.

\textbf{Schur Replay ($\checkmark$), complete generation.} To find the total amount John paid, we need to calculate the total number of shoes and then multiply by the cost of each pair. Step 1: He buys 2 pairs of shoes for each of his 3 children, so the total number of pairs is $2\times3=6$ pairs. Step 2: Each pair costs \$60, so the total cost is $6\times\$60=\$360$. Therefore, the total amount John paid is \$360. The final answer is $\boxed{360}$.

\textbf{Observed difference.} The two baselines change the unit from pairs to individual shoes before applying the \$60 price, doubling the result. Schur Replay preserves the pair-level unit through both arithmetic steps.

\subsection{CMath disagreements}

\paragraph{Qwen3.5-397B-A17B: CMath, question 1085: completed answer versus a truncated reasoning loop.}
The school has 893 students and each uniform costs 88 yuan; the requested estimate is rounded to the nearest ten thousand yuan. The gold answer is 80,000.

\textbf{SOAR ($\checkmark$), CoT excerpt.} ``Total Cost = (Number of Students) $\times$ (Cost per Set). Calculation: $893\times88$. [\emph{calculation omitted}] $78584\to80000$. [\emph{rounding check omitted}] So 80,000 is the correct answer.'' Its final response shows the exact multiplication and emits ``\begin{CJK}{UTF8}{gbsn}8万\end{CJK}'' (or 80000).

\textbf{Four Over Six (incomplete), CoT excerpt.} ``Given: Number of students: 893; Cost per suit: 88 yuan. [\emph{calculation omitted}] Let's do the exact calculation to be sure, then round.'' The trace reaches the same 80,000 conclusion internally, but its archived tail repeats ``Wait, I'll write 80000'' until the generation limit. It contains no closing \texttt{</think>} and no emitted final response.

\textbf{Schur Replay ($\checkmark$), CoT excerpt.} ``$893\times88=78584$. Rounding: $78584\to80000$. Estimation: $900\times90=81000\to80000$. Both match. Final Answer: 80000.'' Its final response gives both exact and estimated calculations and emits $\boxed{80000}$.

\textbf{Observed disagreement.} SOAR and Schur Replay finish their reasoning and emit the gold-equivalent value; Four Over Six enters a repetitive reasoning loop and exhausts the archived generation budget before producing an answer.

\paragraph{Llama-3.3-70B-Instruct.}
\begin{CJK}{UTF8}{gbsn}
\textbf{Question.} 全球荒漠化土地的面积已占陆地总面积的四分之一，而且正在以每年约六万平方千米的速度扩展。地球陆地总面积约为三千六百万平方千米，如果不制止，多少年后地球上将连一根草都看不到？\textbf{ Gold:} 450。

\textbf{SOAR ($\times$), complete generation.} 为了解决这个问题，我们需要计算出荒漠化的速度和全球陆地总面积之间的关系。

1. 全球陆地总面积 = 3600万平方千米。

2. 荒漠化土地面积占陆地总面积的1/4，即荒漠化土地面积 = 3600万 $\times$ 1/4 = 900万平方千米。

3. 每年荒漠化的速度 = 6万平方千米（平均值）。

4. 问题中提到，如果不制止，多少年后地球上连一根草都看不到了，也就是说，所有的陆地都会变成荒漠。

要计算出需要多少年后地球上所有陆地都会变成荒漠，我们需要计算出剩余的陆地面积（非荒漠化的陆地面积）是多少，然后除以每年的荒漠化速度。剩余的陆地面积 = 3600万 $-$ 900万 = 2700万平方千米。年数 = 2700万 / 6万 = 4500年。因此，答案是：$\boxed{4500}$。

\textbf{Four Over Six ($\times$), complete generation.} 为了解决这个问题，我们需要计算荒漠化的速度和全球陆地面积。

1. 全球陆地总面积 = 3600万平方千米。

2. 荒漠化土地面积 = 1/4 $\times$ 3600万 = 900万平方千米。

3. 每年荒漠化的面积 = 6万平方千米。

现在，我们需要计算如果不制止荒漠化，多少年后地球上所有陆地都会变成荒漠。剩余陆地面积 = 全球陆地总面积 $-$ 荒漠化土地面积 = 3600万 $-$ 900万 = 2700万平方千米。由于每年荒漠化的面积为6万平方千米，年数 = 2700万 / 6万 = 4500年。因此，如果不制止荒漠化，4500年后地球上所有陆地都会变成荒漠。最终答案是：$\boxed{4500}$。

\textbf{Schur Replay ($\checkmark$), complete generation.} 要解决这个问题，我们需要按照步骤计算出如果荒漠化以每年6万平方千米的速度扩展，需要多少年后，地球上所有陆地都会变成荒漠。

1. \textbf{计算地球陆地总面积}：已知陆地总面积为3600万平方千米。

2. \textbf{计算当前荒漠化土地面积}：荒漠化土地面积占陆地总面积的1/4，即 $\frac14\times3600\text{万}=900\text{万平方千米}$。

3. \textbf{计算尚未荒漠化的土地面积}：尚未荒漠化的土地面积为总面积减去已荒漠化的面积，即 $3600\text{万}-900\text{万}=2700\text{万平方千米}$。

4. \textbf{计算荒漠化扩展速度}：每年扩展6万平方千米。

5. \textbf{计算需要多少年使所有陆地变成荒漠}：将尚未荒漠化的土地面积除以每年扩展的面积，即 $\frac{2700\text{万}}{6\text{万}}=450$ 年。

因此，如果不采取措施制止荒漠化的扩展，450年后，地球上所有陆地都可能变成荒漠。答案：$\boxed{450}$。
\end{CJK}

\textbf{Observed difference.} All three generations set up the same remaining area and division, but both baselines introduce a factor-of-ten error in the final quotient. Schur Replay evaluates the quotient consistently.

\subsection{LiveCodeBench disagreements}

\paragraph{Qwen3.5-397B-A17B: LiveCodeBench, question 256: direct indegree versus transitive closure.}
The task asks whether partial pairwise ``stronger than'' relations uniquely identify the strongest programmer. Because the archive contains no per-question execution verdict for this Qwen record, we report the program disagreement without assigning correctness marks.

\textbf{SOAR, CoT excerpt.} ``The `strongest programmer' is the one who is stronger than everyone else. [\emph{analysis omitted}] This corresponds to nodes with an in-degree of 0. If there is exactly one such person, they must be the strongest.'' The emitted implementation counts only direct incoming edges:
\begin{lstlisting}[language=Python,basicstyle=\ttfamily\footnotesize,breaklines=true,columns=fullflexible]
in_degree = [0] * (N + 1)
for _ in range(M):
    u = int(next(iterator))
    v = int(next(iterator))
    in_degree[v] += 1
candidates = []
for i in range(1, N + 1):
    if in_degree[i] == 0:
        candidates.append(i)
print(candidates[0] if len(candidates) == 1 else -1)
\end{lstlisting}

\textbf{Four Over Six, CoT excerpt.} ``We need to determine if there is a unique strongest programmer based on the given information. [\emph{analysis omitted}] Find all nodes with in-degree 0; if unique, print it.'' Its emitted implementation likewise uses only direct indegree.

\textbf{Schur Replay, CoT excerpt.} ``The relation `stronger than' is a strict total order. [\emph{analysis omitted}] We need to determine if there is a unique person who must be the strongest programmer consistent with all given information.'' Its emitted implementation first computes reachability:
\begin{lstlisting}[language=Python,basicstyle=\ttfamily\footnotesize,breaklines=true,columns=fullflexible]
for k in range(N):
    for i in range(N):
        for j in range(N):
            if reach[i][k] and reach[k][j]:
                reach[i][j] = True
candidates = [i + 1 for i in range(N)
              if all(not reach[j][i] for j in range(N) if i != j)]
print(candidates[0] if len(candidates) == 1 else -1)
\end{lstlisting}

\textbf{Observed disagreement.} The baselines reason over direct indegree, whereas Schur Replay explicitly materializes the transitive closure before testing whether a unique strongest candidate remains.

\paragraph{Llama-3.3-70B-Instruct.}
\textbf{Problem.} There are $N$ programmers and a strict, total, transitive superiority relation. The input provides $M$ facts $A_i>B_i$. Output the unique programmer that the supplied facts determine to be strongest; output $-1$ if multiple strongest programmers remain possible. Here $2\leq N\leq50$ and $0\leq M\leq N(N-1)/2$. The public examples include a chain that uniquely identifies programmer 1 and a fork for which the answer is $-1$.

\textbf{Acceptance condition.} The complete programs below are executed against the benchmark's public and private stdin/stdout tests. SOAR and Four Over Six fail; Schur Replay passes.

\textbf{SOAR ($\times$), complete generation.}
\begin{lstlisting}[language=Python,basicstyle=\ttfamily\footnotesize,breaklines=true,columns=fullflexible]
from collections import defaultdict, deque
def find_strongest_programmer(n, m, edges):
    graph = defaultdict(list)
    in_degree = [0] * (n + 1)
    for u, v in edges:
        graph[u].append(v)
        in_degree[v] += 1
    queue = deque([i for i in range(1, n + 1)
                   if in_degree[i] == 0])
    while queue:
        u = queue.popleft()
        for v in graph[u]:
            in_degree[v] -= 1
            if in_degree[v] == 0:
                queue.append(v)
    candidates = [i for i in range(1, n + 1)
                  if in_degree[i] == 0]
    if len(candidates) > 1:
        return -1
    else:
        return candidates[0]
n, m = map(int, input().split())
edges = [tuple(map(int, input().split())) for _ in range(m)]
print(find_strongest_programmer(n, m, edges))
\end{lstlisting}
The topological pass mutates every reachable indegree to zero, so the final candidate list no longer identifies the original unique source.

\textbf{Four Over Six ($\times$), complete generation.}
\begin{lstlisting}[language=Python,basicstyle=\ttfamily\footnotesize,breaklines=true,columns=fullflexible]
from collections import defaultdict
def find_strongest_programmer(n, m, relations):
    graph = defaultdict(list)
    for a, b in relations:
        graph[a].append(b)
    def dfs(node, visited):
        if node in visited:
            return False
        visited.add(node)
        for neighbor in graph[node]:
            if not dfs(neighbor, visited):
                return False
        return True
    strongest = None
    for i in range(1, n + 1):
        visited = set()
        if dfs(i, visited) and len(visited) == n:
            if strongest is not None:
                return -1
            strongest = i
    return strongest
n, m = map(int, input().split())
relations = [tuple(map(int, input().split())) for _ in range(m)]
print(find_strongest_programmer(n, m, relations))
\end{lstlisting}
A revisit is treated as failure. Converging paths can therefore reject a valid strongest candidate.

\textbf{Schur Replay ($\checkmark$), complete generation.}
\begin{lstlisting}[language=Python,basicstyle=\ttfamily\footnotesize,breaklines=true,columns=fullflexible]
import sys
def solve():
    N, M = map(int, sys.stdin.readline().split())
    graph = [[False] * N for _ in range(N)]
    for _ in range(M):
        A, B = map(int, sys.stdin.readline().split())
        A -= 1
        B -= 1
        graph[A][B] = True
    for k in range(N):
        for i in range(N):
            for j in range(N):
                if graph[i][k] and graph[k][j]:
                    graph[i][j] = True
    for i in range(N):
        if all(graph[i][j] for j in range(N) if i != j):
            if sum(all(graph[i][j] for j in range(N)
                       if i != j) for i in range(N)) > 1:
                print(-1)
                return
            print(i + 1)
            return
    print(-1)
if __name__ == "__main__":
    solve()
\end{lstlisting}
\textbf{Observed difference.} Schur Replay computes transitive reachability and tests whether a vertex reaches every other vertex, matching the task definition and passing all recovered tests.

\subsection{MGSM-English disagreements}

\paragraph{Qwen3.5-397B-A17B: MGSM-English, question 157: completed arithmetic versus a truncated trace.}
Blake runs back and forth 15 times on a 100-yard field. Kelly runs back and forth once and then to the 40-yard line and back 34 times. The gold difference is 80 yards.

\textbf{SOAR ($\checkmark$), CoT excerpt.} ``One `back and forth' is $100+100=200$ yards. [\emph{steps omitted}] Blake's Distance = $15\times200=3000$ yards. [\emph{steps omitted}] Kelly's total distance is $200+34\times80=2920$ yards.'' The final response emits $\boxed{80}$.

\textbf{Four Over Six (incomplete), CoT excerpt.} ``Blake's Activity: Runs back and forth 15 times. Kelly's Activity: [\emph{problem decomposition omitted}] $15\times200=3000$; $34\times80=2720$; $200+2720=2920$; $3000-2920=80$.'' The archive then continues checking wording and field conventions until its length limit; it contains no closing \texttt{</think>} and no final response.

\textbf{Schur Replay ($\checkmark$), CoT excerpt.} ``One `back and forth' = $100+100=200$ yards. Blake does this 15 times. [\emph{steps omitted}] Kelly's Part 2 is $34\times80=2720$ yards.'' Its final response totals 3,000 and 2,920 yards and emits $\boxed{80}$.

\textbf{Observed disagreement.} SOAR and Schur Replay complete essentially the same correct calculation; Four Over Six reaches 80 inside its CoT but does not exit the reasoning channel or emit a final answer.

\paragraph{Llama-3.3-70B-Instruct.}
\textbf{Question.} Jerry rolls a six-sided die. How much more likely, expressed as a percentage, is rolling a number greater than 3 than rolling two even numbers in a row? Gold: 25.

\textbf{SOAR ($\times$), complete generation.} \textbf{Step 1:} The favorable outcomes greater than 3 are 4, 5, and 6, so the probability is $3/6=1/2$. \textbf{Step 2:} The even outcomes are 2, 4, and 6, so one even roll has probability $3/6=1/2$ and two independent even rolls have probability $(1/2)(1/2)=1/4$. \textbf{Step 3:} The difference is $1/2-1/4=1/4$. \textbf{Step 4:} To express this difference as a percentage, the generation divides it by the probability of two even rolls: $(1/4)/(1/4)\times100=100\%$. \textbf{Step 5:} It interprets this as rolling a number greater than 3 being 100\% more likely, or twice as likely. The final answer is $\boxed{100}$.

\textbf{Four Over Six ($\times$), complete generation.} \textbf{Step 1:} The numbers greater than 3 are 4, 5, and 6, so their probability is $3/6=1/2=0.5$. \textbf{Step 2:} The even numbers are 2, 4, and 6, so the probability of two even rolls is $0.5\times0.5=0.25$. \textbf{Step 3:} The difference is $0.5-0.25=0.25$. \textbf{Step 4:} The generation writes $(0.25/0.25)\times100=100\%$, then repeats $((0.5-0.25)/0.25)\times100=100\%$ as the ``correct percentage increase.'' It concludes that the first event is twice as likely. The final answer is $\boxed{100}$.

\textbf{Schur Replay ($\checkmark$), complete generation.} \textbf{Step 1:} The favorable outcomes greater than 3 are 4, 5, and 6. Out of six outcomes, this gives $3/6=1/2$. \textbf{Step 2:} There are three even outcomes, so two consecutive even rolls have probability $(3/6)(3/6)=9/36=1/4$. \textbf{Step 3:} Subtracting gives $1/2-1/4=1/4$. \textbf{Step 4:} Expressing each probability as a percentage gives $50\%-25\%=25\%$. The final answer is $\boxed{25}\%$.

\textbf{Observed difference.} All methods derive the same event probabilities. The baselines convert their absolute difference into a relative increase, whereas Schur Replay reports the percentage-point difference expected by the benchmark.

\subsection{HumanEval disagreements}

\paragraph{Qwen3.5-397B-A17B: HumanEval, question 22: whether booleans count as integers.}
The task is to filter a heterogeneous Python list, retaining integer values. The canonical body is \texttt{[x for x in values if isinstance(x, int)]}. The disagreement concerns Python's rule that \texttt{bool} is a subclass of \texttt{int}.

\textbf{SOAR, CoT excerpt.} ``In Python, \texttt{bool} is a subclass of \texttt{int}, so \texttt{isinstance(True, int)} returns \texttt{True}. [\emph{deliberation omitted}] To be safe and exclude booleans, I should add \texttt{and not isinstance(value, bool)}.'' It emits:
\begin{lstlisting}[language=Python,basicstyle=\ttfamily\footnotesize,breaklines=true,columns=fullflexible]
return [value for value in values
        if isinstance(value, int) and not isinstance(value, bool)]
\end{lstlisting}

\textbf{Four Over Six, CoT excerpt.} ``Check if the element is an instance of \texttt{int}. Important: In Python, \texttt{bool} is a subclass of \texttt{int}. [\emph{deliberation omitted}] I'll stick with \texttt{isinstance}.'' It emits the canonical rule:
\begin{lstlisting}[language=Python,basicstyle=\ttfamily\footnotesize,breaklines=true,columns=fullflexible]
return [x for x in values if isinstance(x, int)]
\end{lstlisting}

\textbf{Schur Replay (incomplete), CoT excerpt.} ``For each item, check if it is an instance of \texttt{int}. [\emph{deliberation omitted}] What about booleans? In Python, \texttt{bool} is a subclass of \texttt{int}. [\emph{further checking omitted}] I'll stick with \texttt{isinstance}.'' The trace remains in this repeated edge-case analysis until the archive limit and contains no closing \texttt{</think>} or emitted implementation.

\textbf{Observed disagreement.} SOAR deliberately departs from the canonical \texttt{isinstance} semantics by excluding booleans; Four Over Six emits the canonical implementation; Schur Replay reasons toward the canonical rule but does not complete the generation. We report this as a program/completion disagreement rather than inferring a hidden final program.

\paragraph{Llama-3.3-70B-Instruct.}
\textbf{Function requirement.} Implement \texttt{same\_chars(s0, s1)} so that it returns true exactly when two words contain the same distinct characters. The official prompt includes examples with unequal multiplicities, such as \texttt{same\_chars("abcd", "dddddddabc") == True}.

\textbf{SOAR ($\times$), complete implementation.}
\begin{lstlisting}[language=Python,basicstyle=\ttfamily\small,breaklines=true,columns=fullflexible]
def same_chars(s0: str, s1: str) -> bool:
    """Check if two words have the same characters."""
    return sorted(s0) == sorted(s1)
\end{lstlisting}

\textbf{Four Over Six ($\times$), complete implementation.}
\begin{lstlisting}[language=Python,basicstyle=\ttfamily\small,breaklines=true,columns=fullflexible]
def same_chars(s0: str, s1: str) -> bool:
    """Check if two words have the same characters."""
    return sorted(s0) == sorted(s1)
\end{lstlisting}

\textbf{Schur Replay ($\checkmark$), complete implementation.}
\begin{lstlisting}[language=Python,basicstyle=\ttfamily\small,breaklines=true,columns=fullflexible]
def same_chars(s0: str, s1: str) -> bool:
    """Check if two words have the same characters."""
    return set(s0) == set(s1)
\end{lstlisting}

\textbf{Official distinguishing tests.}
\begin{lstlisting}[language=Python,basicstyle=\ttfamily\footnotesize,breaklines=true,columns=fullflexible]
assert candidate('eabcdzzzz', 'dddzzzzzzzddeddabc') == True
assert candidate('abcd', 'dddddddabc') == True
assert candidate('dddddddabc', 'abcd') == True
assert candidate('eabcd', 'dddddddabc') == False
assert candidate('abcd', 'dddddddabcf') == False
assert candidate('eabcdzzzz', 'dddzzzzzzzddddabc') == False
assert candidate('aabb', 'aaccc') == False
\end{lstlisting}
\textbf{Observed difference.} Sorting compares both identity and multiplicity, so both baseline programs fail the positive unequal-multiplicity tests. Schur Replay implements the required set semantics and passes the complete recovered test suite.

\subsection{GPQA-Diamond disagreements}

\paragraph{Qwen3.5-397B-A17B: GPQA-Diamond, question 101: biological regulation versus a reading-frame defect.}
The question asks why a Cre-dependent receptor--eGFP construct shows no green signal. The keyed option is A, a post-excision reading-frame mismatch.

\textbf{SOAR ($\times$), CoT excerpt.} ``The discrepancy lies between the successful protein detection in astrocytes (in vitro) and the lack of fluorescence in neural crest-derived cells (in vivo). [\emph{biological analysis omitted}] Option D suggests they are in a paracrine relationship. [\emph{elimination omitted}] Thus, the paracrine nature of the pair explains why forcing autocrine expression leads to signal loss.'' \textbf{Emitted answer: D.}

\textbf{Four Over Six (incomplete), CoT excerpt.} ``Analyze the Construct Design: [\emph{long construct analysis omitted}] Why would they miss it? Because they relied on the antibody. This narrative supports Option A. [\emph{continued comparison with the paracrine option omitted}]'' The archived trace reaches its length limit while comparing A and D; it contains no closing \texttt{</think>} and no final option.

\textbf{Schur Replay ($\checkmark$), CoT excerpt.} ``After excision, one lox site (34 base pairs) remains at the junction between the Receptor and eGFP sequences. [\emph{derivation omitted}] Since 34 is not divisible by 3, the remaining lox site sequence introduces a 1-base-pair frameshift. [\emph{consequence analysis omitted}] Conclusion: Option A.'' Its final response explains the frameshift and emits A.

\textbf{Observed disagreement.} SOAR selects a biological down-regulation account (D), Four Over Six never commits to a final option, and Schur Replay identifies the construct-level frame defect and emits the keyed A.

\paragraph{Llama-3.3-70B-Instruct.}
\textbf{Question.} Consider
\[
\rho=\frac{1}{2}\left(\lvert0\rangle\!\langle0\rvert+\lvert1\rangle\!\langle1\rvert\right).
\]
What is its geometrical position in the qubit space?

\textbf{Options.} A: $r=(0,0,0)$; B: $r=(1,1,0)$; C: $r=(1,1,1)$; D: $r=(0,0,1)$.

\textbf{Gold rationale.} Since $\lvert0\rangle=(1,0)^\top$ and $\lvert1\rangle=(0,1)^\top$, the state is
\[
\rho=\frac12\begin{pmatrix}1&0\\0&1\end{pmatrix}=\frac12 I.
\]
Comparing this with $\rho=(I+r\cdot\sigma)/2$ gives $r=(0,0,0)$, so the gold option is A.

\textbf{SOAR ($\times$), complete generation:} \texttt{D}

\textbf{Four Over Six ($\times$), complete generation:} \texttt{D.}

\textbf{Schur Replay ($\checkmark$), complete generation.} \textbf{Step 1: Understand the given density matrix.} The given density matrix represents a mixed state where the qubit has an equal probability of being in state $\lvert0\rangle$ and state $\lvert1\rangle$.

\textbf{Step 2: Recall the Bloch sphere representation.} A qubit state can be represented by a point with coordinates $(x,y,z)$ related to the expectation values of the Pauli operators $\sigma_x$, $\sigma_y$, and $\sigma_z$.

\textbf{Step 3: Calculate the expectation values.} For the given density matrix,
\[
\langle\sigma_z\rangle=\frac12\bigl(\langle0\rvert\sigma_z\lvert0\rangle+\langle1\rvert\sigma_z\lvert1\rangle\bigr)=\frac12(1-1)=0.
\]
The expectation values of $\sigma_x$ and $\sigma_y$ are both 0 because the density matrix is diagonal in the computational basis and these operators introduce off-diagonal terms.

\textbf{Step 4: Determine the geometrical position.} Since the expectation values of $\sigma_x$, $\sigma_y$, and $\sigma_z$ are all 0, the point representing this state is at the center of the Bloch sphere. The final answer is $\boxed{A}$.

\textbf{Observed difference.} The two baselines emit the pure-state north-pole distractor D. Schur Replay retains the archived derivation from the density matrix to the vanishing Bloch vector and selects A.

\end{document}